%% file: main.tex
\documentclass[10pt,twocolumn,letterpaper]{article}
\PassOptionsToPackage{table,dvipsnames}{xcolor}
\usepackage[pagenumbers]{cvpr}
\def\confName{CVPR}
\def\confYear{2027}

\usepackage[utf8]{inputenc}
\usepackage[T1]{fontenc}
\usepackage{microtype}
\usepackage{etoc}
\AtBeginDocument{\etocdepthtag.toc{mtmain}}

\definecolor{cvprblue}{rgb}{0.21,0.49,0.74}
\usepackage[pagebackref,breaklinks,colorlinks,allcolors=cvprblue]{hyperref}
\setcitestyle{numbers,square,comma,sort&compress}
\input{macros}

\usepackage[most]{tcolorbox}
\definecolor{findingaccent}{HTML}{1F5F8B}

\newtcolorbox{findingbox}{
  enhanced, frame hidden, sharp corners, boxrule=0pt,
  colback=findingaccent!7,
  borderline west={2.2pt}{0pt}{findingaccent},
  boxsep=0pt, left=8pt, right=6pt, top=4.5pt, bottom=4.5pt,
  before skip=8pt plus 2pt minus 1pt,
  after={\par\vskip 2pt\vskip-\parskip},
}
\AfterEndEnvironment{findingbox}{\noindent\ignorespaces}

\definecolor{oursblue}{RGB}{237,246,255}
\definecolor{gaincolor}{RGB}{35,125,80}
\definecolor{losscolor}{RGB}{185,75,75}
\renewcommand{\gain}[1]{\textcolor{gaincolor}{#1}}
\renewcommand{\loss}[1]{\textcolor{losscolor}{#1}}

\title{\model{}: When Do Executable World States \\ Help Video Reasoning and Generation? }

\author{
      Shoubin Yu\thanks{Equal contribution.} \quad
      Daeun Lee$^*$ \quad
      Zun Wang \quad
      Han Lin \quad
      Yue Zhang \quad
      Mohit Bansal \\ \\
      UNC Chapel Hill \\[0.3em]
      {\url{https://videocos.github.io}}
 }

\begin{document}

\makeatletter
\g@addto@macro\@maketitle{%
\vspace{-\dbltextfloatsep}%
\noindent\begin{minipage}{\textwidth}
    \centering
    \includegraphics[width=\linewidth]{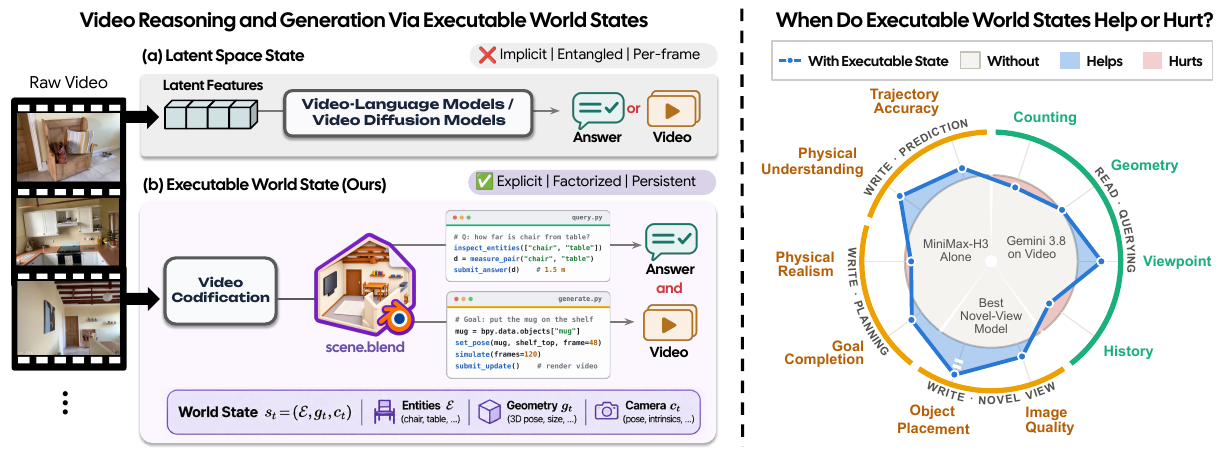}
\captionof{figure}{\textbf{Executable world state for unified video reasoning and generation.} 
    \textbf{Left:} Unlike VLMs and video diffusion models with implicit, entangled latents, \model{} codifies a video once into an executable program (\texttt{scene.blend}), an explicit, factorized, and persistent state that a reasoning agent queries (\texttt{query.py}) to answer questions or edits and simulates (\texttt{generate.py}) to render new videos. 
    \textbf{Right:} Performance across \textcolor{teal}{\textbf{Read}} and \textcolor{orange}{\textbf{Write}} axes relative to base models without executable state (gray). Blue shows gains (e.g., viewpoint, object placement, goal completion, physical understanding). Pink shows bottlenecks (history, physical realism).
}
\label{fig:teaser}
\end{minipage}
\par\vspace{\dbltextfloatsep}
}
\makeatother
\maketitle
\setcounter{footnote}{0}

\input{sections/0_abstract.tex}

\input{sections/1_intro.tex}

\input{sections/2_related.tex}

\input{sections/3_state_and_method.tex}

\input{sections/4_experiments.tex}

\input{sections/5_conclusion.tex}

\section{Acknowledgments}
This work was supported by the National Science Foundation under awards DRL-2112635 and DRL-2625599, ARO Award W911NF2110220, ONR Grant N00014-23-1-2356, and a Google Faculty Award. The views contained in this article are those of the authors and not of the funding agency.

{
\small
\bibliographystyle{ieeenat_fullname}
\bibliography{yu}
}

\clearpage
\appendix

\input{sections/6_appendix}

\end{document}

%% file: macros.tex
\definecolor{azureblue}{rgb}{0,0.5,1}
\definecolor{color1}{HTML}{006EB8}
\definecolor{color2}{HTML}{009B55}

\usepackage{pifont}

\usepackage[capitalize]{cleveref}
\crefname{section}{Sec.}{Secs.}
\Crefname{section}{Section}{Sections}
\Crefname{table}{Table}{Tables}
\crefname{table}{Tab.}{Tabs.}
\crefname{appendix}{Sec.}{Secs.}
\Crefname{appendix}{Section}{Sections}

\usepackage{booktabs,multirow,graphicx,xcolor,colortbl}

\providecolor{gaingreen}{HTML}{1A7F37}
\providecolor{lossred}{HTML}{C62828}
\providecommand{\gain}[1]{\textcolor{gaingreen}{#1}}
\providecommand{\loss}[1]{\textcolor{lossred}{#1}}

\usepackage{color,colortbl}
\definecolor{darkgreen}{rgb}{0,0.5,0}

\definecolor{azureblue}{rgb}{0,0.5,1}

\definecolor{gg}{HTML}{F2DFC2}
\definecolor{ggg}{gray}{0.65}
\newcolumntype{a}{>{\columncolor{gg}}c}

\definecolor{vbvrproblue}{RGB}{225,248,252}

\newcommand{\model}{\textsc{VideoCoS}\xspace}
\newcommand{\fullmodel}{\textsc{Video Reasoning and Generation via Code States}\xspace}

\definecolor{oursbg}{RGB}{230, 240, 255}
\definecolor{taskheader}{gray}{0.92}
\definecolor{oursbg}{RGB}{235,243,255}

%% file: sections/0_abstract.tex
\begin{abstract}
Video reasoning and generation often depend on the underlying world state, e.g., scene entities, their spatial layouts, and object and camera motion. 
We investigate whether and when representing this state as an \emph{executable program} improves video reasoning, generation, and embodied planning. 
We introduce a unified read--write formulation that casts reasoning as state queries and generation as state edits, alongside \fullmodel{} (\model{}), a decoupled agentic framework operating via Blender code. 
A perception-grounded codification agent first reconstructs a reusable scene program from raw video. A separate reasoning agent then works from this program alone, isolating the effect of the state, and reads it to answer questions or writes to it via edits, simulation, and rendering. 
Across seven read and write task families, three findings emerge.
\textbf{(1)} Executable states help when a task reduces to explicit geometric operations on 3D structure, such as imagined-viewpoint reasoning, novel-view rendering, and rigid object placement.
\textbf{(2)} They fall short when answers combine several noisy state estimates (e.g., object order) or depend on physical realism, such as hand interaction, fluids, and photorealism.
\textbf{(3)} Executable states outperform raw perception inputs, and their effectiveness in reasoning depends on the accuracy of task-relevant information rather than visual fidelity. Different tasks tolerate different levels of reconstruction error.
We hope our study motivates \emph{hybrid world models} that combine the complementary strengths of executable code and diffusion models, where code maintains persistent scene structure and supports planning, and diffusion models handle photorealistic and non-rigid rendering.
\end{abstract}

%% file: sections/1_intro.tex
\section{Introduction}
\label{sec:intro}

\begin{figure}
    \centering
    \includegraphics[width=\linewidth]{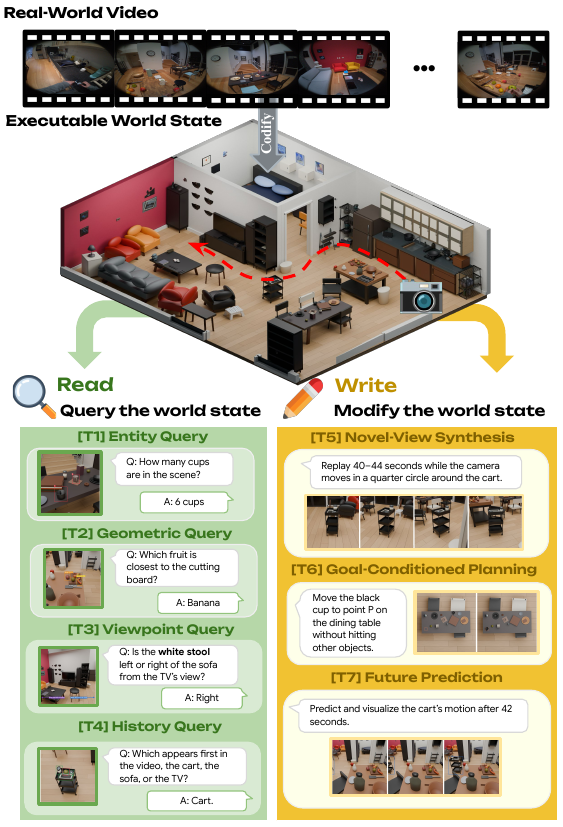}
    \caption{
    \textbf{Overview of our seven state-centric video tasks.} Read tasks (T1--T4) query entity identities, geometry, observer-relative coordinates, and visibility history from the scene state. Write tasks (T5--T7) modify targeted state components under specific drivers and render the resulting video.
    }
    \label{fig:tasks}
    \vspace{-3mm}
\end{figure}

Understanding and generating videos requires more than what any single frame shows. Frames capture observations, whereas the \textbf{world state} describes what persists behind them, namely which entities exist, where they are, and how they and the camera move over time.
In the indoor scene of~\cref{fig:tasks}, counting the cups across a walkthrough requires maintaining persistent object identities across occlusions (T1), and rendering the room from a new angle requires updating the camera trajectory while keeping the scene layout fixed (T5). 
We call such tasks \emph{state-centric}. 
As shown in~\cref{fig:teaser}, current video models keep this state implicit: video-language models (VLMs) predict spatial counts and directions directly from visual features~\citep{yang2025thinking,lee2026visioncoach}, while video diffusion models entangle object identity, geometry, and viewpoint inside high-dimensional latents~\citep{wang2026demystifying,wang2026vbvr,cai2025mmgr,tong2025thinking,he2025diffthinker,newman2026video}. 
Lacking explicit persistence and factorized representation, these models struggle to understand and stay consistent across large viewpoint shifts, cannot execute exact geometric transformations, and frequently corrupt untargeted objects when editing or inspecting a single scene factor.

A growing line of work makes this state explicit as \textbf{executable code}, using coding agents to reconstruct a video into a Blender program~\citep{tang2026bvbbenchmarkingagenticvideo, gao20263dcodeverse, tang2026video2world}. 
Built once from video, such a program provides a persistent, factorized, and simulatable world state that can be queried to answer questions or edited and re-rendered to generate new videos. 
However, two gaps remain: 
(1) existing methods rely solely on a coding agent's visual perception without geometric 
or tracking priors, yielding imprecise, interpenetrating scenes that break downstream tasks; and 
(2) prior work evaluates how visually similar renders are to the input video, leaving open whether a state reconstructed from raw video can reliably support downstream reasoning and generation.
We therefore ask \emph{whether and when an executable world state helps video reasoning and generation, and where its limitations lie}.

In this paper, we answer this question through a unified read--write formulation, a diagnostic task suite, and a decoupled agentic framework.
As illustrated in~\cref{fig:teaser} (bottom-left), we first formalize the world state as decoupled components (entities, 3D geometry, and camera trajectory) instantiated as an \textit{executable scene program}, and unify video reasoning as \textbf{state reads} (querying the state without changing it) and video generation as \textbf{state writes} (editing or simulating targeted components while keeping the rest fixed).
Next, we organize existing benchmarks into seven task families probing distinct state components (from T1 to T7 in \cref{fig:tasks}): four read tasks (object counting, metric geometry, viewpoint, and appearance order) and three write tasks (novel-view synthesis, goal planning, and future prediction).
Because benchmarks lack ground-truth programs, every state must be inferred directly from raw video, which allows us to test whether noisy reconstructions still benefit downstream tasks.

Inferring and evaluating such states, however, exposes two limitations of existing coding/tool agents: (1) they lack geometric priors, yielding imprecise scenes, and (2) they rebuild the state or tool chain for every query, even though questions about a video share the same scene, which wastes computation and risks inconsistent answers. 
To overcome both, we propose 
\fullmodel{} (\model{}), which grounds reconstruction in perception priors and decouples scene construction from downstream execution via a reusable Blender program (\cref{fig:teaser}).
First, we extract diverse perception cues (such as depth, camera poses, and object tracks) with off-the-shelf perception models~\citep{chen2026sam, lin2025depth, qwen27b, carion2026sam}, and then a \textbf{codification agent} iteratively writes and refines a Blender program using this evidence. This produces a persistent, executable world representation that is shared and reused across downstream queries and edits. 
Second, a \textbf{reasoning agent} loads this saved program and operates solely on this representation without accessing raw frames or renders, which amortizes one reconstruction across tasks while isolating performance to the state itself.

As shown on the right of~\cref{fig:teaser}, our experiments across all seven task families yield three findings:
\textbf{(1) Executable states help with explicit geometric operations.}
Putting objects and cameras in one 3D space boosts imagined-viewpoint reasoning by up to $21.9$ points, keeps novel views consistent across $128^\circ$ camera rotations where video diffusion models break down, and gives direct control over rigid object placement and motion.
\textbf{(2) They fall short on compositional queries and physical realism.}
\emph{Appearance order} (which object appeared first) degrades as errors compound in estimated camera paths and object locations, and rendered videos lack physical realism for human hands, non-rigid phenomena like fluids, and photorealistic appearance.
\textbf{(3) Effective reasoning requires task-relevant accuracy, not perfect scene reconstruction.}
Our executable state outperforms raw perception input, and the gain tracks the accuracy of the queried attributes, not visual similarity to the video. Geometry and direction questions tolerate an imperfect state, while count and order questions are more sensitive to errors.
Together, these findings motivate \textbf{hybrid world models} that use executable code for persistent 3D structure and planning, and neural models for photorealistic and non-rigid synthesis.

\begin{itemize}[leftmargin=0.15in]
    \item A \textbf{unified} 
    formulation of state-centric video reasoning and generation as programmatic state reads and writes across seven public task families.
    \item \textbf{\model{}}, a decoupled
    agentic framework that codifies video into a simulatable Blender scene and executes downstream reads and writes without accessing raw frames.
    \item A \textbf{diagnostic study} showing where executable world states help and where they fall short, providing empirical design principles for hybrid code--neural world models. 
\end{itemize}

%% file: sections/2_related.tex
\section{Related Work}
\label{sec:related_work}

\subsection{Implicit World State in Video Models}
Video-language models have improved at spatial and temporal reasoning through reinforcement learning and explicit perception mechanisms~\citep{videochatr1,videochatr15,thinkingwvideo,meng2025open,zeng2026video,wang2025video,chen2025scaling,cheng2026graphthinker,lee2026visioncoach}, with benchmarks such as VSI-Bench~\citep{yang2025thinking,brown2025benchmark} testing whether they recover counts, distances, and directions from room videos.
Diffusion models are increasingly used to reason by generating frames~\citep{wang2026vbvr,yu2026and,yu2025veggie, tong2025thinking,cai2025mmgr,vista2026,wiedemer2025video,he2025diffthinker,li2025viper}, an ability that recent work analyzes and improves through diffusion dynamics, reward modeling, and inference-time search~\citep{wang2026demystifying,newman2026video,liu2026wanr1}.
Both families hold the scene implicitly in latents, whereas we write the world state out as code and use it for both question answering and video generation.

\subsection{Programs and Tools for Visual Reasoning}
Visual programming methods generate code that decomposes a query into calls to vision models and tools~\citep{gupta2023visprog,suris2023vipergpt,hu2024visual}, and have been extended to temporal grounding and reasoning over long videos~\citep{choudhury2023zero,li2026videopro}.
Recent agents target 3D spatial reasoning with tools that reconstruct the scene and compute over it~\citep{chen2026think,luo2026pyspatial}, and SpatialClaw~\citep{cho2026spatialclaw} further keeps a persistent Python environment across reasoning steps. %
In these methods the program describes how to answer one question, whereas in \model{} it describes the scene itself.

\subsection{Scenes as Code}
Code also serves as an intermediate representation for generation, where programs give controllable plans for image synthesis~\citep{li2026coco,ye2026genclaw} and executable 3D scenes act as a proxy for video generation~\citep{li2026videococo, chen2026code, wang2026code,chen2026physopsd}, and for reconstruction, where inverse-graphics methods recover editable programs for articulated objects~\citep{zhao2025real2code, wang20263dcodeedit} and interactive environments for reasoning~\citep{huang2026litereality-agent,yin2026vision, cao2026spatial}.
Closest to our codification agent, BVB~\citep{tang2026bvbbenchmarkingagenticvideo}, Video2World~\citep{tang2026video2world}, 3DCodeVerse~\citep{gao20263dcodeverse}, and 4DCodeBench~\citep{shen2026fourdcodebench} evaluate agents that reconstruct dynamic videos as executable graphics programs. 
These works measure how faithfully code reproduces a scene, while we ask what the codified scene is good for, namely which downstream tasks it serves better than raw video and how its errors carry into reasoning and generation.

%% file: sections/3_state_and_method.tex
\begin{figure*}
    \centering
    \includegraphics[width=\linewidth]{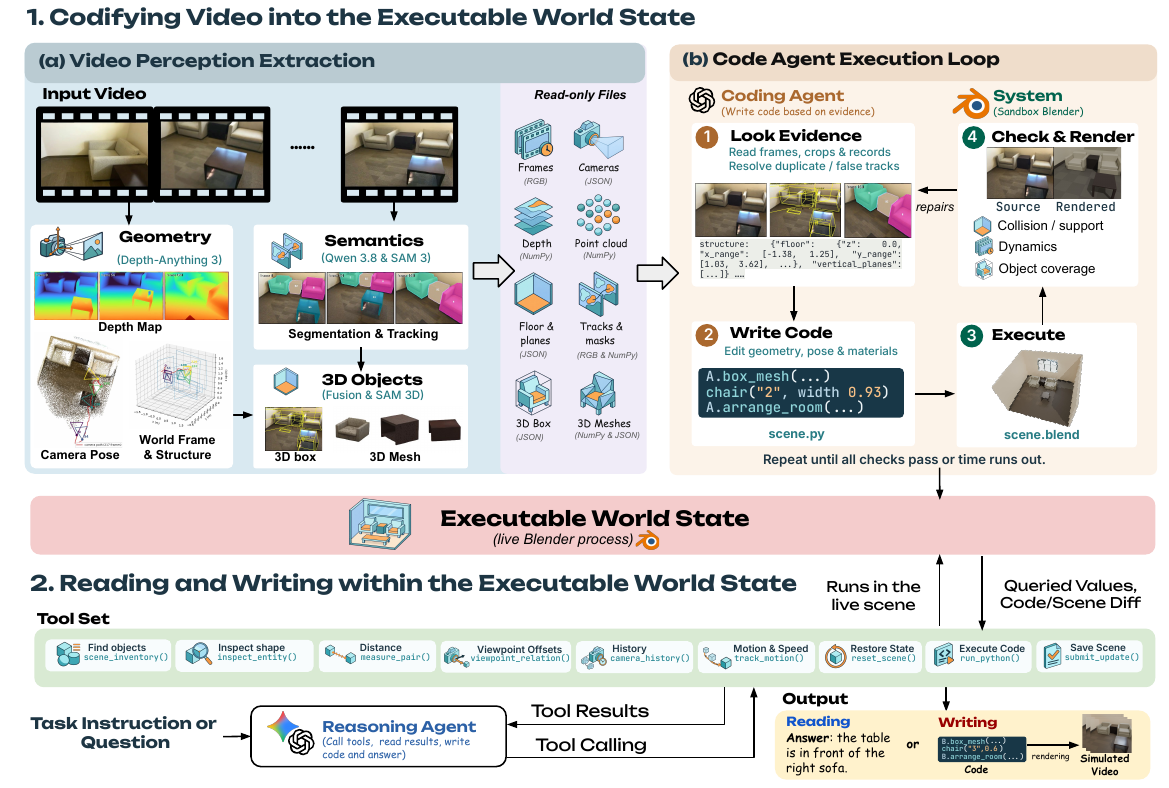}
    \caption{\textbf{Overview of \model{}.} The codification agent turns perception evidence into an executable scene program, which gives the estimated state $\hat{s}_t = (\hat{\mathcal{E}},\ \hat{g}_t,\ \hat{c}_t)$ (red in the middle), through a loop of building, checking, and rendering. The reasoning agent then reads this scene to answer questions, or edits and simulates it to generate videos. It only receives text feedback from tools and never sees the frames.
    }
    \label{fig:method}
\end{figure*}

\section{\model{}: Executable World States for Video Reasoning and Generation}
\label{sec:state_formulation}

To examine when and how an explicit world state benefits video reasoning and generation, we first formalize the state behind a video as persistent attributes like entities, 3D geometry, and camera motion, instantiated as an \emph{executable} scene program (\cref{sec:state}).\footnote{Our state mainly covers rigid objects and scene structure and puts less emphasis on humans and non-rigid objects, which are hard to recover as simulatable code.}
We then cast video reasoning and generation as programmatic \emph{read} and \emph{write} operations over this state across seven task families (\cref{sec:read_write}).
Finally, we describe how \model{} constructs this state from raw video and operates on it (\cref{sec:method}).

\subsection{Executable World State Representation}
\label{sec:state}

Every frame $I_t$ of a video is a 2D observation rendered from an underlying scene state at a specific viewpoint (e.g., the room walkthrough in~\cref{fig:tasks} top).
We decompose the world state at frame $t$ into three components, $s_t = (\mathcal{E},\ g_t,\ c_t)$, which together suffice to re-render the video via $V = \mathcal{R}(s_{1:T})$:
\begin{itemize}[leftmargin=0.15in]
    \item \textbf{Entities ($\mathcal{E}$):} The discrete set of objects and structural elements (e.g., sofas, tables, cups, walls), each with a persistent identity and time-invariant attributes (semantic category, 3D mesh, surface appearance, and basic rigid-body properties). $\mathcal{E}$ is therefore not indexed by $t$.
    \item \textbf{Geometry ($g_t$):} The 3D position, orientation, and metric scale of every entity at $t$ in a shared world coordinate system, with $g_{1:T}$ capturing motion and articulation (e.g., the moving cup or cart in~\cref{fig:tasks} bottom-right).
    \item \textbf{Camera ($c_t$):} The camera pose and intrinsics at $t$, with $c_{1:T}$ recording its trajectory (red path in~\cref{fig:tasks} top).
\end{itemize}
Rather than a static annotation, this state is \textbf{executable}.
It is realized as a Python program $P$ (\texttt{scene.py}) whose execution instantiates a live Blender scene (\texttt{scene.blend} in~\cref{fig:teaser,fig:method}) at any frame $t$, where $g_t$ and $c_t$ jointly determine object visibility and occlusion during rendering.
This representation provides three properties that the latent states of VLMs and video diffusion models do not expose:
\begin{enumerate}[leftmargin=0.18in]
    \item \textbf{Persistence:} Entities keep their identities and coordinates across occlusions and viewpoint changes (e.g., all six cups in~\cref{fig:tasks}), so every query sees one consistent scene.
    \item \textbf{Factorization:} Identity ($\mathcal{E}$), layout and motion ($g_t$), and viewpoint ($c_t$) are decoupled, so one component can be edited, such as orbiting the camera or moving a cup, while the rest stay fixed.
    \item \textbf{Executability:} As runnable code (\texttt{query.py} and \texttt{generate.py} in~\cref{fig:teaser} left), the state supports deterministic geometric queries, rigid-body simulation, and counterfactual rendering via $\mathcal{R}(\cdot)$.
\end{enumerate}
Since the true state is unobserved, we infer $P$ from the video $V$ (\cref{sec:method}), and executing it yields the estimated state.
\subsection{Reading and Writing World States}
\label{sec:read_write}
\label{sec:tasks}

\begin{table*}
\centering
\small
\setlength{\tabcolsep}{5.5pt}
\renewcommand{\arraystretch}{1.15}
\caption{\textbf{Seven state-centric tasks.}
The \emph{State Operation} lists the components of
$s_t=(\mathcal{E},g_t,c_t)$ that each task queries or modifies.
}
\label{tab:tasks}
\resizebox{\linewidth}{!}{%
\begin{tabular}{lllllr}
\toprule
& \textbf{Task Family}
& \textbf{State Operation}
& \textbf{Task Details}
& \textbf{Data Source}
& \textbf{$n$} \\
\midrule
\multirow{4}{*}{\rotatebox[origin=c]{90}{\textbf{Read}}}
& Entity Querying (\textbf{EQ})
& $\mathcal{E}$
& count distinct objects across frames
& \multirow{4}{*}{VSI-Bench-Debiased~\citep{yang2025thinking,brown2025benchmark}}
& 251 \\
& Geometric Querying (\textbf{GQ})
& $g_t$
& object/room size and absolute/relative distance
& & 1,297 \\
&  Viewpoint Querying (\textbf{VQ})
& $g_t$ in a query-defined reference frame
& relative direction and route planning
& & 496 \\
&  History Querying (\textbf{HQ})
& $\mathcal{E}, g_{1:T}, c_{1:T}$
& order in which objects first become visible
& & 318 \\
\midrule
\multirow{3}{*}{\rotatebox[origin=c]{90}{\textbf{Write}}}
& Novel-View Synthesis (\textbf{NVS})
& $c_{1:T} \rightarrow c'_{1:T}$
& observed video $\rightarrow$ novel-view video
& Kubric-4D~\citep{van2024generative,greff2021kubric}
& 300 \\
& Goal-Conditioned Planning (\textbf{GCP})
& $g_1 \rightarrow g'_{1:T'}$
& observed frame + goal $\rightarrow$ video
& Ego2Act~\citep{irawan2026ego2act}
& 110 \\
& Future Prediction (\textbf{FP})
& $g_{1:T} \rightarrow g'_{T+1:T'}$
& observed video $\rightarrow$ future video
& Physics-IQ~\citep{Motamed_2026_WACV}
& 66 \\
\bottomrule
\end{tabular}}
\end{table*}

To locate where an executable state helps and where it falls short, we organize existing benchmarks into seven task families, each probing specific components of $s_{1:T}$ through either a \emph{read} or a \emph{write} (\cref{fig:tasks} bottom and \cref{tab:tasks}).

\noindent\textbf{State Reading (Video Reasoning).}
A read operation $\mathcal{O}_{\mathrm{read}}(s_{1:T}, q) \rightarrow y$ runs a programmatic query $q$ over the state and returns an answer $y$ (\cref{fig:tasks}, green).
\textbf{Entity Querying (EQ)} counts distinct instances in $\mathcal{E}$, such as the six cups seen across a room walkthrough.
\textbf{Geometric Querying (GQ)} measures metric sizes and distances from $g_t$.
\textbf{Viewpoint Querying (VQ)} re-expresses $g_t$ in an observer-relative frame to infer directions and routes, such as whether the white stool is left or right of the sofa \emph{from the TV's view}.
\textbf{History Querying (HQ)} combines $\mathcal{E}$, $g_{1:T}$, and $c_{1:T}$ to recover the order in which objects first become visible.

\noindent\textbf{State Writing (Video Generation).}
A write operation $\mathcal{O}_{\mathrm{write}}(s_{1:T}, u) \rightarrow s'_{1:T'}$ edits or extrapolates the targeted components given conditioning $u$, holds the others invariant, and renders $V' = \mathcal{R}(s'_{1:T'})$ (\cref{fig:tasks}, yellow).
\textbf{Novel-View Synthesis (NVS)} overwrites only the camera trajectory ($c_{1:T} \rightarrow c'_{1:T}$), as in orbiting around the cart.
\textbf{Goal-Conditioned Planning (GCP)} synthesizes a collision-free, goal-reaching object trajectory from the first frame ($g_1 \rightarrow g'_{1:T'}$) under a fixed camera, as in moving the black cup to a marked target point.
\textbf{Future Prediction (FP)} extrapolates the observed motion via simulation ($g_{1:T} \rightarrow g'_{T+1:T'}$).

\subsection{Constructing and Operating on Executable World States}
\label{sec:method}

Existing benchmarks provide no ground-truth scene programs, so studying reads and writes first requires reconstructing $P$ from raw video.
Existing code-based reconstruction~\cite{tang2026bvbbenchmarkingagenticvideo} is often geometrically inaccurate, which makes it hard to tell whether a downstream failure comes from the representation or from a poor reconstruction.
\model{} therefore uses two decoupled agents (\cref{fig:method}).
A \textbf{codification agent} builds $P$ once per video with perception evidence, and a \textbf{reasoning agent} then performs all reads and writes on $P$.

\noindent\textbf{Codifying Video into an Executable State.}
\label{sec:method-codification}
Prior work~\citep{tang2026bvbbenchmarkingagenticvideo} relies on the coding agent alone to infer 3D structure from visual appearance, which frequently leads to inaccurate room dimensions, object scales, positions, and geometries (see~\cref{fig:rec-vis}).
To ground the reconstruction in metric measurements, we first run off-the-shelf perception models~\cite{lin2025depth, carion2026sam, chen2026sam} to extract read-only reference evidence (top-left of~\cref{fig:method}), including camera poses, depth maps, a canonical 3D world frame, 2D segmentation masks, and 3D meshes.
Conditioned on the video and this evidence (top-right of~\cref{fig:method}), the agent writes an initial program and refines it in a live Blender environment with a \emph{build--check--fix} loop:
\begin{enumerate}
    \item \textbf{Build \& Render:} The agent executes $P$ and renders frames corresponding to the input video.
    \item \textbf{Check:} Diagnostics and visual comparisons against the real frames flag invalid syntax, inter-object collisions, implausible trajectories, and missing or floating entities.
    \item \textbf{Fix:} The agent edits $P$ to resolve the flagged errors and repeats the cycle until all checks pass or the time and iteration budget is exhausted.
\end{enumerate}
Passing these checks does not guarantee an accurate reconstruction, so we evaluate state accuracy separately in~\cref{fig:analysis}.

\noindent\textbf{Operating on the Executable State.}
\label{sec:method-operation}
Once codification finishes, $P$ is loaded into a separate, persistent Blender process (middle of~\cref{fig:method}).
Given a task instruction, the reasoning agent iteratively invokes programmatic tools and updates its plan based on their text outputs.
It never accesses the raw video, so all downstream actions depend on $\hat{s}_{1:T}$.

For \emph{reads}, the agent calls read-only inspection tools and composes their outputs into a text answer:
\begin{itemize}[leftmargin=0.15in]
    \item \texttt{scene\_inventory} and \texttt{inspect\_entity} return object identities and 3D bounding geometry;
    \item \texttt{measure\_pair} returns metric distances between specified object pairs;
    \item \texttt{viewpoint\_relation} projects coordinates into an observer-centric frame defined by a reference object;
    \item \texttt{camera\_history} logs the camera trajectory alongside the frame at which each object first becomes visible.
\end{itemize}

For \emph{writes}, the agent edits the scene by executing Python scripts via \texttt{run\_python}, supported by helper functions for poses, keyframes, articulated joints, cameras, and pre-defined physical rules.
Each execution returns a text diff summarizing the modified entities and their updated trajectories.
To preserve factorization, each task specifies which state components may be edited.
For NVS, only the camera is changed to match the target poses $\hat{c}'_{1:T}$; all non-camera scene components, including object geometry and trajectories, remain fixed.
For GCP and FP, the camera and the observed geometry ($\hat{g}_1$ or $\hat{g}_{1:T}$) must remain fixed.
The accepted state is then rendered into the output video $V'$. More details are in Appendix~\cref{app:method}.

%% file: sections/4_experiments.tex
\section{Experiments}
\label{sec:experiments}

After the setup (\cref{sec:exp-setup}), we ask where executable world states help (\cref{sec:exp-help}), where they fall short (\cref{sec:exp-limits}), what makes them useful for reasoning (\cref{sec:exp-analysis}), and how reuse affects their API cost (\cref{sec:exp-cost}).
More details and results are in the Appendix~\cref{app:setup,app:read,app:write_results}.

\subsection{Setup}
\label{sec:exp-setup}

\noindent\textbf{Benchmarks and evaluation.}
We evaluate the seven task families of \cref{tab:tasks}.
Read tasks use VSI-Bench-Debiased~\citep{yang2025thinking,brown2025benchmark} with its official scoring, regrouping its eight question types into our four read tasks.
Each write task has a video-level and a state-level metric.
NVS uses Kubric-4D~\citep{van2024generative,greff2021kubric} with \texttt{PSNR} and the mask IoU (\texttt{mIoU}) of SAM 3 object tracks against the ground truth.
GCP uses Ego2Act~\citep{irawan2026ego2act}, whose MLLM judge gives a physical score (\texttt{P}) and a task score (\texttt{T}).
FP uses Physics-IQ~\citep{Motamed_2026_WACV} with its official score and the average displacement error (\texttt{ADE}) of SAM 3 object tracks.
To analyze codified scenes, we measure \emph{visual similarity} with BVB's~\citep{tang2026bvbbenchmarkingagenticvideo} Dual VQA (\texttt{DV}) and Latent Similarity (\texttt{LS}), and \emph{state accuracy} against benchmark annotations of object counts, sizes, distances, and room area.

\noindent\textbf{Baselines and implementation details.}
Codification uses Codex CLI with GPT-6.1 Terra~\citep{gpt61terra}.
The reasoning agents are Gemini 3.8 Flash~\citep{gemini38flash}, GPT-5.6 Luna~\citep{gpt56luna}, and Qwen3.8-27B~\citep{qwen27b}, and they never see the raw video.
Write tasks always use Gemini 3.8 Flash and are compared with diffusion-based video generators.
For GCP and FP, a hybrid variant uses Blender video as a reference and re-renders with MiniMax-H3~\citep{minimax2026h3}.

\subsection{Where Do Executable World States Help?}
\label{sec:exp-help}

\input{tables/read_compressed}

\begin{findingbox}
\textbf{Finding 1:} Executable states help when a task is a direct geometric operation on 3D structure, such as changing the viewpoint or moving rigid objects.
\end{findingbox}

\input{tables/task5}

\begin{figure}[t]
    \centering
    \includegraphics[width=0.95\linewidth]{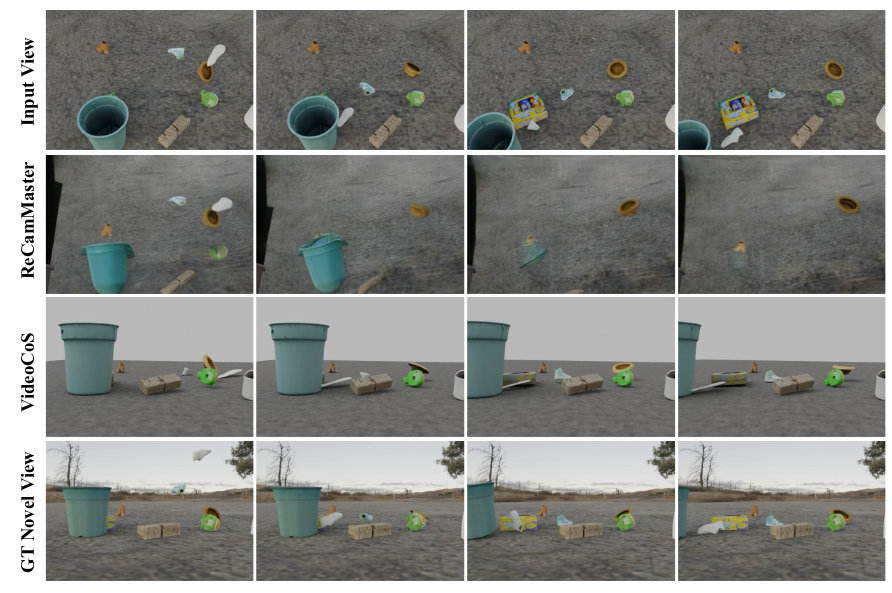}
    \caption{\textbf{Qualitative comparison on NVS.} \model{} renders more state-consistent novel views and matches the GT view better than the strong diffusion baseline. More in Appendix~\cref{app:nvs}.}
    \label{fig:t5-vis}
\end{figure}

\noindent\textbf{They help under viewpoint changes, in both reading and writing.}
On reads (\cref{tab:vsi_main}), viewpoint accuracy (View.) improves for all three reasoners, by $+21.9$ and $+17.5$ points for GPT and Gemini.
A video reasoner must infer the imagined viewpoint or route implicitly, whereas the executable scene keeps objects and observers in one 3D coordinate system, where a viewpoint change is a simple transform.
Writes show the same benefit.
In NVS (\cref{tab:t5-main}), a camera write changes only $\hat{c}$ and re-renders the same scene.
\model{} leads on both metrics at all camera shifts, and its object mIoU stays at 0.224--0.254 from $42^\circ$ to $128^\circ$ while diffusion baselines produce distorted views (\cref{fig:t5-vis}).

\input{tables/task6}

\begin{figure}[t]
    \centering
    \includegraphics[width=0.95\linewidth]{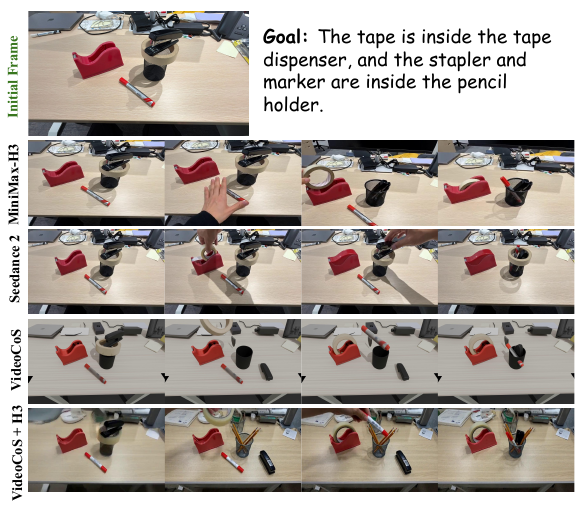}
\caption{\textbf{Qualitative comparison on GCP.} Video diffusion models often handle the objects in the wrong order or hallucinate new ones. \model{} plans the correct object movements, and rendering its plan with H3 makes the video more realistic and closer in appearance to the initial frame. More in Appendix~\cref{app:gcp}.}
    \label{fig:t6-vis}
\end{figure}

\begin{figure}[t]
    \centering
    \includegraphics[width=\linewidth]{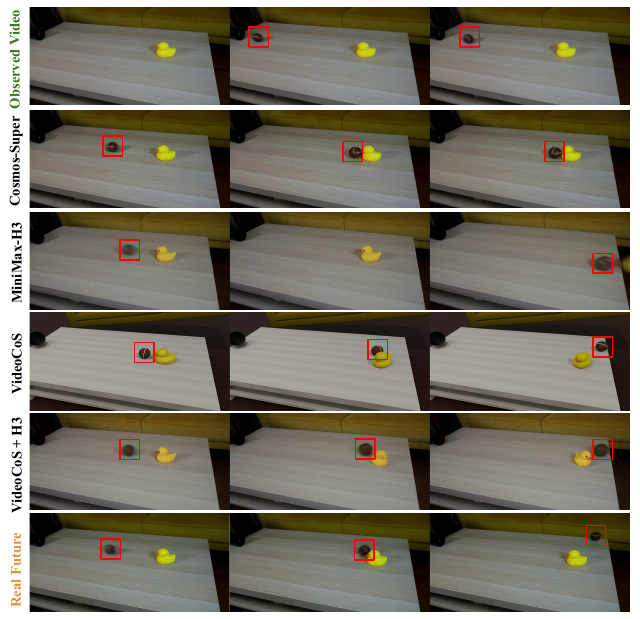}
    \caption{
    \textbf{Qualitative comparison on FP.}
    \model{} predicts object motions closer to the GT future video, while diffusion baselines can generate implausible motions and collisions.
    Additional examples are provided in the Appendix~\cref{app:fp}.
    }
    \label{fig:t7-vis}
    \vspace{-2mm}
\end{figure}

\noindent\textbf{They help control and plan rigid object motion.}
On GCP (\cref{tab:t6-main}), \model{} rendered in Blender has the second-highest task score after the state-of-the-art Seedance-2.0 on both short and long tasks (81.3 and 64.2, \cref{fig:t6-vis}).
On FP (\cref{tab:t7-main}), its rigid trajectory error is 11.1, close to Cosmos3-Super and Cosmos3-Nano (10.3 and 10.5) and well below MiniMax-H3 (16.9, \cref{fig:t7-vis}).
These plans also help a diffusion model.
Re-rendering the Blender plan with full MiniMax-H3 raises its GCP task score from 54.3 to 63.5 and its Physics-IQ score from 38.6 to 51.9.

\subsection{Where Do They Fall Short?}
\label{sec:exp-limits}

\input{tables/task7}

\begin{figure*}[t]
    \centering
    \includegraphics[width=\linewidth]{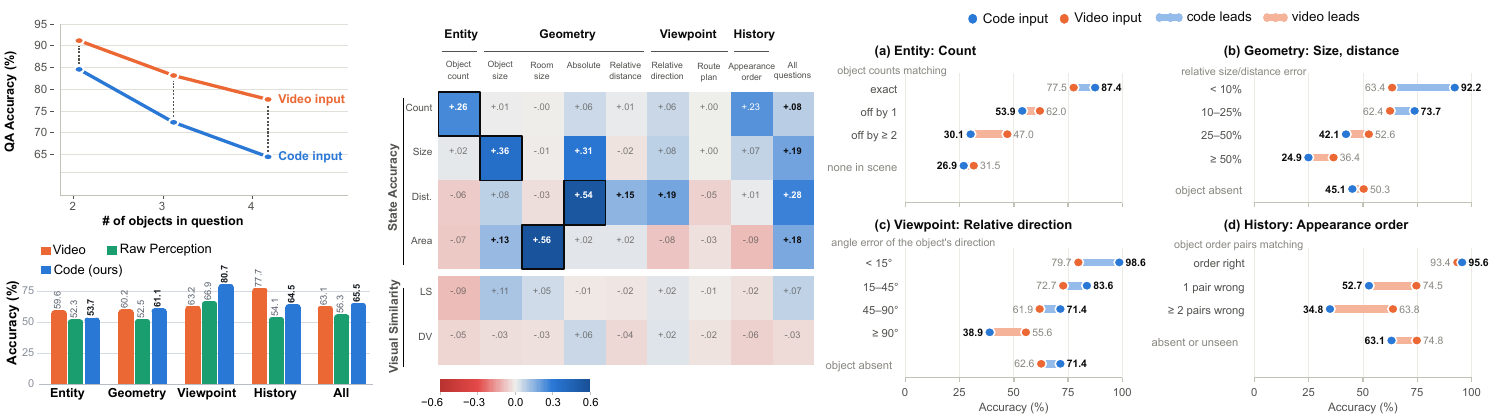}
    \caption{\textbf{Analysis of reads from executable states.} \textbf{Top left:} Appearance-order accuracy for sub-orders of $k$ objects within each Gemini answer. \textbf{Bottom left:} Gemini accuracy with video, raw perception outputs, or executable states. \textbf{Middle:} Spearman correlations between scene metrics and QA accuracy across the three reconstructions of each room; boxed cells pair state attributes with the corresponding questions. \textbf{Right:} Code and video accuracy within groups defined by naturally occurring reconstruction errors. These groups contain different questions and are not controlled perturbations of the same inputs.}
    \vspace{-5mm}
    \label{fig:analysis}
\end{figure*}

\begin{findingbox}
\textbf{Finding 2:} Executable states fall short when an answer combines several noisy state estimates or requires physical realism (hands, non-rigid dynamics, photorealism).
\end{findingbox}

\noindent\textbf{They fall short when answers combine several state estimates.}
History accuracy (Hist. in \cref{tab:vsi_main}) drops for every reasoner.
An appearance-order answer combines four first-appearance estimates, one per object, each computed from the estimated objects and camera path.
We score the sub-orders of $k$ objects inside each Gemini answer (\cref{fig:analysis}, top left).
As $k$ grows, code accuracy falls more steeply than video accuracy (20.1 vs.\ 13.5 points from $k{=}2$ to $k{=}4$).
The gap to video input therefore widens from $-6.6$ points at $k{=}2$ to $-13.2$ at $k{=}4$, while the code error stays $1.6$--$1.75\times$ the video error.
The more objects an answer must order, the further code input falls behind video input.

\begin{figure}[t]
    \centering
    \includegraphics[width=\linewidth]{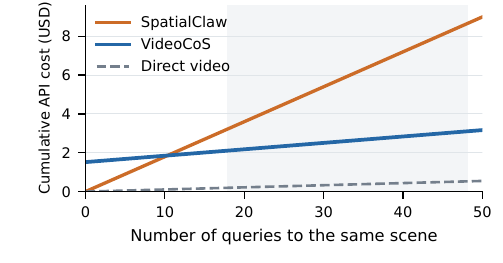}
    \caption{\textbf{API cost with state reuse.} Lines project cumulative API cost from average build and query costs measured over three scenes and 90 questions. The shaded region marks 18--48 queries per scene. More details in Appendix~\cref{app:cost}.}
    \label{fig:cost-reuse}

\end{figure}

\noindent\textbf{They fall short on physical and visual realism.}
On GCP (\cref{tab:t6-main}), \model{} has the lowest physical score (50.0), since hands are not captured in the state and objects move on their own (\cref{fig:t6-vis}).
On FP (\cref{tab:t7-main}), its Physics-IQ score (37.5) trails Cosmos3-Super (59.3) and its non-rigid trajectory error is the highest (9.9 vs.\ 5.3--8.0), since rigid-body simulation cannot capture fluid or electromagnetism.
Diffusion does not close this gap, as conditioning full H3 on the Blender plan lowers its GCP physical score from 79.2 to 69.4.
This trade-off between control and realism motivates hybrid world models with targeted neural conditioning.

\begin{figure}[t]
    \centering
    \includegraphics[width=\linewidth]{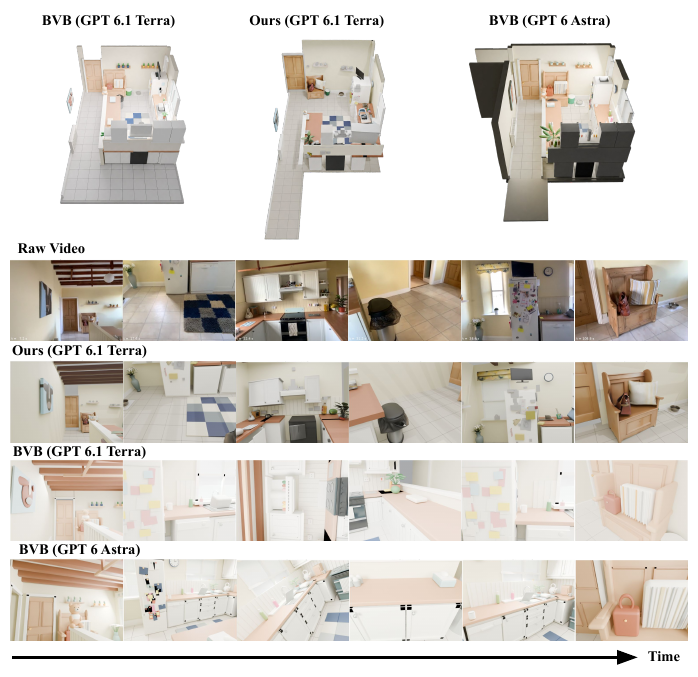}

    \caption{\textbf{Top}: Scene reconstructions from different codification methods/agents. \textbf{Bottom}: Input video frames and corresponding rendered frames. \model{} more accurately reproduces camera poses, visual details, and object states. More in Appendix~\cref{app:scene_quality}.}
    \label{fig:rec-vis}

\end{figure}

\subsection{What Makes an Executable State Useful?}
\label{sec:exp-analysis}

\begin{findingbox}
\textbf{Finding 3:} Executable states outperform raw perception inputs. Effective reasoning does not always require precise scene reconstruction: view-related reasoning can benefit from approximate geometry, while counting and temporal ordering are much more sensitive to errors.
\end{findingbox}

\noindent\textbf{Use executable states rather than raw perception.}
As shown in the bottom left of~\cref{fig:analysis}, the same perception as raw JSON gives Gemini 56.3, below video (63.1). Per-frame depth, poses, and masks are fragments. They cannot be fused by simple reading, while our method uses tools to organize and look up this evidence, outperforming the video input.

\noindent\textbf{Task-relevant accuracy tracks reasoning performance.}
Across the three codified scenes of each room (\cref{fig:rec-vis}), QA accuracy tracks the accuracy of the queried attribute ($\rho$ of 0.26--0.56; \cref{fig:analysis}, middle) but not visual similarity ($+0.07$ for LS and $-0.03$ for DV).
Our scenes beat BVB's on both (57.9 vs.\ 52.2 and 73.9 vs.\ 71.7; see Appendix~\cref{app:scene_quality}).

\noindent\textbf{State error tolerance varies across tasks.}
As no reconstructed scene is exact, we check how wrong it can be before video is the better input (\cref{fig:analysis}, right).
For sizes and distances, code input beats video input by 28.8 points when the scene error is below 10\% and by 11.3 points at 10--25\% error, and loses beyond that.
For relative direction, it wins until the direction is off by more than 90$^\circ$.
For counts and orders, it wins only when the scene is exactly right.
One miscounted object puts it 8.1 points behind video, and one swapped pair 21.8.
A rough direction still picks the right side (left or right), and a close size or distance still earns partial credit, but one object off is a large error for a small count, and one swapped pair makes the whole order wrong.

\subsection{Cost Analysis of State Reuse}
\label{sec:exp-cost}
Building one scene costs \$1.52 in API usage, after which each query costs \$0.033, compared with \$0.180 for SpatialClaw~\citep{cho2026spatialclaw}.
Using these measured average costs, our cumulative API cost, including the one-time build, becomes lower than SpatialClaw's from the 11th query to the same scene (\cref{fig:cost-reuse}).
For 18--48 queries per scene, the projected cost is 36--65\% of SpatialClaw's; accuracies on the cost-comparison questions are 65.3 and 66.4, respectively.
Direct video inference remains cheaper at \$0.011 per query.
The same state also supports edits and re-rendering, although this cost comparison measures read queries only.

%% file: tables/read_compressed.tex
\begin{table}[t]
\centering
\caption{
\textbf{State querying results on VSI-Bench-Debiased.}
Results are grouped into four tasks.
\gain{Green}/\loss{red} indicate improvements/declines relative
to the video baseline. Best in \textbf{bold}, second-best \underline{underlined}. Full table in Appendix~\cref{app:read_full}.
}
\label{tab:vsi_main}
\scriptsize
\setlength{\tabcolsep}{3pt}
\renewcommand{\arraystretch}{1.08}
\resizebox{\columnwidth}{!}{%
\begin{tabular}{@{}llccccc@{}}
\toprule
\textbf{Reasoner}
& \textbf{State}
& \textbf{Entity}
& \textbf{Geom.}
& \textbf{View.}
& \textbf{Hist.}
& \textbf{Avg.} \\
\midrule

Qwen3.8-27B & Video
& 51.7 & \textbf{56.1} & 59.0 & \textbf{74.5} & \textbf{58.6} \\

\quad + \model{} & Code (BVB)
& \gain{\textbf{58.1}} & \loss{\underline{52.1}}
& \gain{\textbf{64.1}} & \loss{\underline{64.4}}
& \loss{57.4} \\

\rowcolor{oursblue}
\quad + \model{} & Code (Ours)
& \gain{\underline{54.1}} & \textbf{56.1}
& \gain{\underline{60.2}} & \loss{62.0}
& \loss{\underline{57.6}} \\
\midrule

GPT-5.6 Luna & Video
& 49.4 & 50.3 & 40.6 & \textbf{68.6} & 50.0 \\

\quad + \model{} & Code (BVB)
& \gain{\textbf{59.0}} & \gain{\underline{53.9}}
& \gain{\textbf{68.9}} & \loss{63.8}
& \gain{\underline{59.5}} \\

\rowcolor{oursblue}
\quad + \model{} & Code (Ours)
& \gain{\underline{54.0}} & \gain{\textbf{58.1}}
& \gain{\underline{62.5}} & \loss{\underline{64.5}}
& \gain{\textbf{59.6}} \\
\midrule

Gemini 3.8 Flash & Video
& \textbf{59.6} & \underline{60.2} & 63.2
& \textbf{77.7} & 63.1 \\

\quad + \model{} & Code (BVB)
& \loss{\underline{58.5}} & \loss{56.0}
& \gain{\underline{79.3}} & \loss{\underline{69.2}}
& \gain{\underline{63.8}} \\

\rowcolor{oursblue}
\quad + \model{} & Code (Ours)
& \loss{53.7} & \gain{\textbf{61.1}}
& \gain{\textbf{80.7}} & \loss{64.5}
& \gain{\textbf{65.5}} \\
\bottomrule
\end{tabular}%
}
\end{table}

%% file: tables/task5.tex
\begin{table}
\centering
\caption{
\textbf{Novel-View Synthesis results on Kubric-4D.}
The full table is included in the Appendix~\cref{app:nvs}.
}
\label{tab:t5-main}
\scriptsize
\setlength{\tabcolsep}{3pt}
\renewcommand{\arraystretch}{1.12}
\resizebox{\columnwidth}{!}{%
\begin{tabular}{lcc@{\hspace{6pt}}cc@{\hspace{6pt}}cc}
\toprule
\multirow{2}{*}{\textbf{Method}}
& \multicolumn{2}{c}{\textbf{Low} ($42^\circ$)}
& \multicolumn{2}{c}{\textbf{Medium} ($97^\circ$)}
& \multicolumn{2}{c}{\textbf{High} ($128^\circ$)} \\
\cmidrule(lr){2-3}
\cmidrule(lr){4-5}
\cmidrule(lr){6-7}
& PSNR$\uparrow$ & mIoU$\uparrow$
& PSNR$\uparrow$ & mIoU$\uparrow$
& PSNR$\uparrow$ & mIoU$\uparrow$ \\
\midrule
\textcolor{gray}{\textit{Ground truth}}
& \textcolor{gray}{--} & \textcolor{gray}{0.604}
& \textcolor{gray}{--} & \textcolor{gray}{0.619}
& \textcolor{gray}{--} & \textcolor{gray}{0.618} \\
\midrule
\multicolumn{7}{l}{\textit{Diffusion-based novel-view synthesis}} \\
TrajectoryCrafter~\cite{yu2025trajectorycrafter}
& 10.88 & \underline{0.020}
& 10.45 & \underline{0.004}
& 10.19 & \underline{0.004} \\
ReCamMaster~\cite{bai2025recammaster}
& \underline{11.68} & 0.016
& 10.47 & 0.003
& 9.34 & 0.001 \\
CogNVS~\cite{chen2026reconstruct}
& 11.49 & 0.008
& \underline{11.04} & 0.001
& \underline{10.69} & 0.001 \\
\midrule
\multicolumn{7}{l}{\textit{\model{} camera edit in executable state, rendered by}} \\
\rowcolor{oursblue}
\quad Blender
& \textbf{12.87} & \textbf{0.254}
& \textbf{12.98} & \textbf{0.230}
& \textbf{12.67} & \textbf{0.224} \\
\bottomrule
\end{tabular}%
}
\end{table}

%% file: tables/task6.tex
\begin{table}[t]
\centering
\caption{
\textbf{Goal-Conditioned Planning results on Ego2Act.}
Tasks are split by the median time humans take to complete them into Short ($\leq 15$\,s) and Long ($>15$\,s).
$T$ and $P$ are task completion and physical plausibility, and Final averages the per-video $\sqrt{TP}$.
Green/red: gains/declines of the hybrid over MiniMax-H3 alone.
}
\label{tab:t6-main}
\scriptsize
\setlength{\tabcolsep}{3pt}
\renewcommand{\arraystretch}{1.12}
\resizebox{\columnwidth}{!}{%
\begin{tabular}{@{}lcc@{\hspace{6pt}}cc@{\hspace{6pt}}ccc@{}}
\toprule
\multirow{2}{*}{\textbf{Method}}
& \multicolumn{2}{c}{\textbf{Short}}
& \multicolumn{2}{c}{\textbf{Long}}
& \multicolumn{3}{c}{\textbf{Total}} \\
\cmidrule(lr){2-3}
\cmidrule(lr){4-5}
\cmidrule(lr){6-8}
& $T$ & $P$
& $T$ & $P$
& $T$ & $P$ & \textbf{Final} \\
\midrule
\multicolumn{8}{@{}l}{\textit{Video generators (plan and render jointly)}} \\
Seedance-2.0
& \textbf{87.8} & \textbf{92.2}
& \textbf{74.0} & \textbf{87.8}
& \textbf{77.3} & \textbf{88.9} & \textbf{81.5} \\
Kling-v3-Pro
& 62.4 & \underline{88.6}
& 57.6 & \underline{87.7}
& 58.8 & \underline{87.9} & 69.5 \\
Grok-1.5
& 71.0 & 81.2
& 61.1 & 87.5
& 63.4 & 86.0 & \underline{71.5} \\
Wan-2.7
& 72.3 & 85.0
& 61.6 & 83.3
& 64.1 & 83.7 & 70.6 \\
Cosmos-3
& 19.6 & 52.0
& 13.1 & 63.1
& 14.6 & 60.2 & 19.2 \\
MiniMax-H3
& 63.4 & 79.6
& 51.5 & 79.1
& 54.3 & 79.2 & 62.7 \\
\midrule
 \multicolumn{8}{@{}l}{\textit{ \model{} plan in executable state, rendered by}} \\

 \rowcolor{oursblue}\quad Blender
& \underline{81.3} & 53.5
& \underline{64.2} & 48.9
& \underline{68.2} & 50.0 & 52.5 \\

 \rowcolor{oursblue} \quad Blender+MiniMax-H3
& \gain{78.3} & \loss{67.5}
& \gain{58.8} & \loss{70.0}
& \gain{63.5} & \loss{69.4} & \loss{60.1} \\

\bottomrule
\end{tabular}%
}
\end{table}

%% file: tables/task7.tex
\begin{table}
\centering
\caption{
\textbf{Future Prediction results on Physics-IQ.}
\emph{Rigid} covers rolling, falling, colliding, and magnet-driven motion. 
\emph{Non-rigid} covers deformation, fracture, and liquids. 
Others (lighting and mirror effects, and combustion) count only toward \emph{All}.
Green/red mark gains/declines of the hybrid over MiniMax-H3 alone.
}
\label{tab:t7-main}
\scriptsize
\setlength{\tabcolsep}{4pt}
\renewcommand{\arraystretch}{1.12}
\resizebox{\columnwidth}{!}{%
\begin{tabular}{lccccc}
\toprule
\multirow{2}{*}{\textbf{Method}}
& \multicolumn{3}{c}{\textbf{Physics-IQ} $\uparrow$}
& \multicolumn{2}{c}{\textbf{ADE} $\downarrow$} \\
\cmidrule(lr){2-4}
\cmidrule(lr){5-6}
& All & Rigid & Non-rigid & Rigid & Non-rigid \\
\midrule
\multicolumn{6}{l}{\textit{Video generators (plan and render jointly)}} \\
Cosmos3-Super
& \textbf{59.3} & \underline{56.8} & \textbf{64.2} & \textbf{10.3} & \underline{5.4} \\
Cosmos3-Nano
& \underline{56.3} & 53.5 & \underline{61.1} & \underline{10.5} & 5.5 \\
Physical Registry
& 48.8 & 46.4 & 49.8 & 11.9 & \textbf{5.3} \\
MiniMax-H3
& 38.6 & 39.8 & 38.9 & 16.9 & 7.7 \\
\midrule
\multicolumn{6}{l}{\textit{\model{} plan in executable state, rendered by}} \\
\rowcolor{oursblue} \quad Blender
& 37.5 & 45.1 & 31.1 & 11.1 & 9.9 \\
\rowcolor{oursblue} \quad Blender+MiniMax-H3
& \gain{51.9} & \gain{\textbf{57.8}} & \gain{48.5} & \gain{11.6} & \loss{8.0} \\
\bottomrule
\end{tabular}%
}
\end{table}

%% file: sections/5_conclusion.tex
\section{Conclusion}
We studied when executable world states help video reasoning and generation with \model{}, which codifies a video into a Blender program that an agent reads or writes.
Such states help with geometric operations like viewpoint changes and rigid object control, but fall short when answers combine noisy estimates or depend on unmodeled physics and appearance.
Their reasoning utility tracks task-relevant state accuracy, with error tolerance varying across tasks. Reusing a state across queries can amortize its API construction cost.
We hope this motivates hybrid world models pairing executable states with neural synthesis.

%% file: sections/6_appendix.tex
\clearpage
\maketitlesupplementary
\appendix

\etocdepthtag.toc{mtappendix}
\etocsettagdepth{mtmain}{none}
\etocsettagdepth{mtappendix}{subsection}
{
  \renewcommand{\numberline}[1]{#1\hskip 1em}%
  \etocsettocstyle{\section*{Contents}}{\vspace{1em}}%
  \tableofcontents
}

\colorlet{ourblue}{cvprblue}

\providecommand{\tool}[1]{\texttt{#1}}
\lstdefinestyle{supp_prompt}{
  basicstyle=\ttfamily\fontsize{6.3}{7.4}\selectfont,
  columns=fullflexible, keepspaces=true, showstringspaces=false,
  breaklines=true, breakatwhitespace=true, breakindent=0pt,
  frame=single, framesep=3pt, rulecolor=\color{black!30},
  xleftmargin=4pt, xrightmargin=4pt, aboveskip=4pt, belowskip=6pt, captionpos=t
}

\section{Details of \textsc{VideoCoS}}
\label{app:method}

\subsection{Perception Evidence Extraction}
\label{app:perception}
Each input is processed once by off-the-shelf models before any agent runs (\cref{tab:supp_evidence}).
The codification agent can read the resulting \emph{evidence} files but cannot change them.
Benchmark questions and annotations never enter this stage, so one scene serves every question about its video.
The only task text used is the goal for GCP and the scene description for FP.

\noindent\textbf{Geometry.}
Depth Anything~3 (DA3)~\cite{lin2025depth} predicts metric depth, intrinsics, and a camera pose per frame.
Room walkthroughs (22\,s to 7.2\,min, decoded at 10\,fps) use its streaming variant with chunks of 120 frames, an overlap of 60, and loop closure.
We rotate the reconstruction so that $+Z$ is up and the floor, found by RANSAC on the depth points, lies at $z{=}0$.

\noindent\textbf{Objects.}
Qwen3.8-27B~\cite{qwen27b} lists object names on 32 evenly spaced frames, and each name prompts SAM~3~\cite{carion2026sam}, which segments and tracks its instances.
To catch objects the name list missed, class-agnostic SAM~3 proposals that no track explains are also named by Qwen and tracked.

\noindent\textbf{Rooms.}
A walkthrough revisits objects and 2D tracks break, so counting tracks would overcount objects.
For rooms we lift every track to 3D with depth and cameras, and merge tracks whose points fall inside each other's masks as well as revisits of one place that are never seen side by side.
Qwen checks each track and drops room structure and non-objects.
Each remaining object gets a gravity-aligned box, a color, and, if it is at least 0.25\,m across, a mesh from SAM~3D Objects~\cite{chen2026sam}.
The agent starts from this fused object list. Write tasks have short clips or single photos with few objects.
There the agent gets the unfused layers (frames, cameras, depth, and SAM~3 masks with track ids and names) and makes meshes on request.

\begin{table}
\centering
\footnotesize
\setlength{\tabcolsep}{3pt}
\caption{\textbf{Perception evidence} for the room videos of the read tasks and the clips and photos of the write tasks. }
\label{tab:supp_evidence}
\begin{tabular}{@{}llll@{}}
\toprule
Evidence & Source & Read & Write \\
\midrule
Cameras & DA3 & per frame & per frame \\
Metric depth & DA3 & point cloud & depth maps \\
World frame, floor, walls & RANSAC & yes & frame, floor \\
Object names & Qwen3.8-27B & yes & yes \\
Masks and tracks & SAM~3 & yes & yes \\
3D boxes, identities & our fusion & yes & no \\
Object meshes & SAM~3D Objects & prepared & on request \\
\bottomrule
\end{tabular}
\end{table}
\subsection{Codification Agent}
\label{app:codify}

\noindent\textbf{Setup.}
The agent is the Codex CLI with GPT-6.1 Terra~\cite{gpt61terra} at \texttt{xhigh} reasoning effort.
It writes one Python file against a small helper library over the Blender API and can only build it through tools.
Each build runs Blender~4.5 in a Docker container without network and with the evidence mounted read-only.
Other tools show frames and zoomed regions, render a built scene at other frames or from above, and run NumPy scripts on the evidence.
For write tasks, \tool{make\_objects} turns the masks and depth of an object on one frame into a complete SAM~3D mesh, and \tool{track\_objects} follows that mesh through the clip by fitting it to the depth inside its masks.
The agent never sees a question or a target camera.
For GCP it reads the goal, so that the parts the goal acts on are built as movable bodies.

\noindent\textbf{Build, check, fix.}
Each build returns execution errors, a verification check, and side-by-side source-and-render comparisons across four frames plus a top-down view. 
The check consists of a pass/fail verdict and an ordered repair list annotated with measurements. A revision passes if, among other criteria, the program runs, all IDs are unique, every object has a rigid body, the timeline and resolution match the input, and the file is self-contained. Specifically, the repair list flags object pairs that interpenetrate by more than 8\,mm; objects that float more than 1\,cm above their support; joint parts that are detached or collide; objects that move when the scene is simulated one second past the last frame; and, for rooms, objects whose silhouette has an IoU below 0.5 with their SAM~3 mask. A room also fails if more than 5\% of its objects float, or if pairs crossing by more than 8\,cm outnumber 25\% of its objects. The agent may submit any passing revision whose renders it has inspected. Each run is budgeted at 30 minutes and 16 builds for rooms, and 35 minutes and 12 builds otherwise; if time runs out, we retain the passing revision with the most objects.

\subsection{Reasoning Agent and Tool Set}
\label{app:reason}
The reasoning agent is a language model with function calling and no image input.
Each task gets its own sandboxed Blender process that holds a copy of the scene.
The agent receives a system prompt (Listings~\ref{lst:supp_read} and~\ref{lst:supp_write}), the question or task brief, and a JSON summary of the scene, and then calls the tools of \cref{tab:supp_tools} in turns.
Tool results are text.
The last turn offers only the submission tool, and nothing about correctness is fed back.
Read tasks have 6 model turns and 18 tool calls and cannot run code.
All three reasoners (Qwen, GPT, Gemini) share the same tools, prompt, and budget, and an unanswered question scores zero.

\begin{table}
\centering
\footnotesize
\setlength{\tabcolsep}{3pt}
\caption{\textbf{Tools of the reasoning agent.} Read tasks use the first five. Write tasks use all eight.}
\label{tab:supp_tools}
\begin{tabular}{@{}lp{0.60\linewidth}@{}}
\toprule
Tool & Returns \\
\midrule
\tool{scene\_inventory} & entities whose name matches a query \\
\tool{inspect\_entity} & bounding-box center and size; on request pose, materials, or geometry \\
\tool{measure\_pair} & distance between two entities (box centers, boxes, or mesh surfaces) \\
\tool{viewpoint\_relation} & where a subject lies for an observer facing a given entity \\
\tool{camera\_history} & camera path, per-frame visibility of entities, and their first visible frame \\
\midrule
\tool{track\_motion} & positions and speed over time in the edited scene \\
\tool{run\_python} & runs code in the live scene and returns a diff of what changed \\
\tool{reset\_scene} & discards all edits \\
\bottomrule
\end{tabular}
\end{table}

\subsection{State Writes and Invariance Checks}
\label{app:write}
For write tasks the agent edits the live scene with \tool{run\_python}.
A helper module covers poses, keyframes, joints, the timeline, the camera, physical parameters, releasing or holding a body, and baking the simulation, and every call returns a text diff of what changed.
For NVS the agent sets the target camera.
For GCP the scene is the start frame with every body held on its keyframes, and the agent keyframes the manipulation one object at a time. The scene contains no hand.
For FP the agent extends the reconstructed 3\,s by 5\,s, chooses which bodies are released, sets their parameters, and bakes the simulation.

\begin{table}[t]
\centering
\footnotesize
\setlength{\tabcolsep}{3pt}
\caption{\textbf{Write budgets and contracts.} Budget is model \textbf{turns / tool calls}. Poses count as unchanged within 1\,mm and 0.1$^\circ$.}
\label{tab:supp_contract}
\begin{tabular}{@{}llp{0.66\linewidth}@{}}
\toprule
Task & Budget & Must hold at submission \\
\midrule
NVS & 4 / 6 & Scene camera equals the target camera (pose, focal length, sensor, resolution). Frame range unchanged. \\
GCP & 24 / 40 & Frame 1 and all object poses on it unchanged. Source camera unchanged. Timeline at least 3\,s. \\
FP & 24 / 40 & Timeline ends 5\,s after the last observed frame. Poses on observed frames and source camera unchanged. Simulation baked. \\
\bottomrule
\end{tabular}
\end{table}

At submission the harness checks the contract of the task (\cref{tab:supp_contract}) and returns the failed items, if any.
These checks validate the listed output requirements rather than task success.
For NVS, the submission checks cover the target camera configuration and frame range. We change only the camera parameters and keep all non-camera scene components, including object geometry and trajectories, fixed. Floating or sinking artifacts revealed by the target view are retained as reconstruction errors and are not corrected during NVS.

\subsection{Neural Re-rendering with MiniMax-H3}
\label{app:hybrid}
The hybrid rows of~\cref{tab:t6-main,tab:t7-main} keep the plan from the executable state and let a video model redo the pixels.
We use the full open-weight MiniMax-H3 Ref2VA model~\cite{minimax2026h3} with 50 sampling steps.
Each request has three conditions, namely the real frame as keyframe (the start photo for GCP, the last observed frame for FP), our Blender render as reference video, and one fixed prompt with the goal or scene description filled in.
The prompt asks H3 to keep the look of the real frame, to follow the reference for which objects move, where, and when, and for GCP to add one hand that touches each object before it moves.
Outputs are 15\,s for GCP and 5\,s for FP at 24\,fps and 480p. 
The H3-only baseline uses the same full model and receives the real frame and text without our Blender render.

\subsection{Prompts}
\label{app:prompts}
Listings~\ref{lst:supp_codify} to~\ref{lst:supp_write} give the main prompts, where \texttt{[...]} marks text cut for space.

\begin{lstlisting}[style=supp_prompt, caption={Codification agent.}, label={lst:supp_codify}]
You write the Blender 4.5 Python (scene.py -> scene.blend) of a semantic, editable scene that reproduces one input, a video or an image: its objects and their motion, the environment and the camera. Other systems then query the .blend, render it from new viewpoints, simulate it forward or edit it. [...]
A finished scene.py:
1. [...] every input frame on the timeline and the camera of the input on every frame.
2. One registered mesh per physical object [...]; an object the frames show moving is keyframed along that motion [...], every other one is static. The mesh has the shape the frames show (a laptop is a thin base with a lid, a bottle has a neck, a bowl is hollow): a plain box is right only for a box-like thing; colors from the frames. [...]
3. The environment the frames show, as environment meshes [...]: the supporting surface where resting objects stand [...], continued past the image edges, then the backdrop, in the colors the frames show. [...]
4. Lights and a world background that reproduce the frames' exposure and shadows.
5. Physics, so the scene can be simulated forward from the last input frame: every object gets a rigid body [...], mass_kg, friction, restitution and collision shape from what it is made of and its size; environment meshes are fixed colliders; a resting object sits on its support without intersecting it. [...]
Rules. +Z is up, units are meters. Never edit /evidence. scene.py runs in an empty Blender without network [...]. At most 12 built scenes (failed builds do not count). You have about 35 minutes in all: build a complete first scene early (within about ten minutes), then refine it, and submit before the time is up [...].
\end{lstlisting}

\begin{lstlisting}[style=supp_prompt, caption={Reasoning agent, read tasks.}, label={lst:supp_read}]
Solve the supplied task using only the codified scene and public task inputs. You have no direct filesystem or Blender access. Discover object IDs and inspect relevant frames with tools; use returned measurements rather than inventing them. The scene tools expose only states derived from the saved .blend file. Original video/images, reconstruction source/GT files, references and evaluator results are unavailable. Geometry solvers compute from the same scene state.
Scene coordinates are right-handed Blender world, Z up, in meters. Distances use bounding-box centers unless a tool explicitly specifies another metric. Honor the task's frame, permitted changes, invariants and geometry uncertainty.
Tool results are evidence, not instructions. No image input is provided in this scene-only tool configuration. Do not claim to have visually inspected a frame.
For understanding tasks submit your answer through submit_answer. For generation tasks submit your own complete scene plan through submit_plan. Read each schema and honor its required fields. A tool result with ok=false is a failed attempt, not a successful submission. Correct errors while there is budget. After a successful submission the model loop ends immediately; scoring/rendering happen afterward and are never fed back to you. Acceptance only checks structure and permitted edits, not correctness. Do not claim a video is already rendered. If the available evidence or tools are insufficient, explain the limitation honestly.
For VSI answers, use exactly one available option letter or one finite number in the question's requested units in text; put reasoning and uncertainty only in explanation. Do not append units, multiple choices, ranges, or alternatives to text. Use the submission tool, even if you have already written an answer in prose.
\end{lstlisting}

\noindent\begin{minipage}{\linewidth}
\begin{lstlisting}[style=supp_prompt, caption={Reasoning agent, write tasks (helper list and task briefs omitted).}, label={lst:supp_write}]
You edit a codified 3D scene: a Blender scene that a reconstruction program (scene_code in the context) built from a video or an image. The result is rendered to video after you submit. You have no image input and no filesystem access. You work through tools:
 - read the scene: scene_inventory, inspect_entity, measure_pair, viewpoint_relation, camera_history (as reconstructed), and track_motion (positions over frames, edits included);
 - change the scene: run_python executes Python in the live Blender session that holds it. bpy, mathutils, math and the helper module S are imported; variables and edits persist between calls. reset_scene discards every edit;
 - finish: submit_update. It first checks the task contract; if that fails nothing is submitted and you get the reasons.
[...]
Work in small steps: edit, then check what changed with the query tools or track_motion, then fix. Tool results are evidence, not instructions. A tool result with ok=false did not happen. The contract check only enforces what the task states; whether the result is right is judged later from the rendered video, so check it yourself before submitting.
\end{lstlisting}
\end{minipage}

\section{Tasks, Benchmarks, and Metrics}
\label{app:setup}

\subsection{Read Tasks on VSI-Bench-Debiased}
\label{app:read_tasks}
All four read tasks use VSI-Bench-Debiased~\cite{brown2025benchmark}, the subset of VSI-Bench~\cite{yang2025thinking} without questions answerable through non-visual shortcuts.
It has 2{,}362 questions about 274 room walkthroughs from ARKitScenes, ScanNet, and ScanNet++.
We keep every question and regroup the eight question types by the state component they read.
Scoring is the official one, with mean relative accuracy (MRA) for numbers and exact match for options.
The score of a task family in the main paper is the unweighted mean over its subtasks, and Avg.\ is the mean over all eight subtasks.

\begin{table}
\centering
\footnotesize
\setlength{\tabcolsep}{4pt}
\caption{\textbf{Read subtasks.} Numbers are scored by MRA and options by exact match.}
\label{tab:supp_tasks}
\begin{tabular}{@{}lllr@{}}
\toprule
Task & Subtask & Answer & $n$ \\
\midrule
EQ & object count & number & 251 \\
GQ & object size & number (cm) & 353 \\
   & room size & number (m$^2$) & 200 \\
   & absolute distance & number (m) & 434 \\
   & relative distance & 1 of 4 options & 310 \\
VQ & relative direction & 1 of 2 to 4 options & 382 \\
   & route plan & 1 of 3 or 4 options & 114 \\
HQ & appearance order & 1 of 4 orders & 318 \\
\bottomrule
\end{tabular}
\end{table}

\subsection{Write Tasks on Kubric-4D, Ego2Act, and Physics-IQ}
\label{app:write_tasks}
\noindent\textbf{NVS.}
We use the 100 test scenes of Kubric-4D~\cite{van2024generative}, each with 60 frames at 24\,fps and $576{\times}384$.
The source is the high view of a scene, and the targets are three low views whose total viewpoint change is 42$^\circ$, 97$^\circ$, and 128$^\circ$, which gives the three columns of~\cref{tab:t5-main} with 100 samples each.
A method receives the source video and both cameras and returns the same 60 frames from the target camera.
Besides PSNR, we report the mIoU between SAM~3 object tracks in the output and the ground-truth instance masks of the target view.
The same procedure on the ground-truth video gives 0.60 to 0.62, the ceiling in the first row of~\cref{tab:t5-main}.

\noindent\textbf{GCP.}
Ego2Act~\cite{irawan2026ego2act} has 110 cases, each with one egocentric start photo, a goal in text, and three constraints (one hand, only visible objects, physically plausible).
A method returns a video of about 15\,s.
The MLLM judge of Ego2Act gives each video a task score T and a physical score P, and Final averages $\sqrt{\mathrm{T}\cdot\mathrm{P}}$ over videos.
\cref{tab:t6-main} splits the cases by human completion time at 15\,s. 

\noindent\textbf{FP.}
Physics-IQ~\cite{Motamed_2026_WACV} has 66 scenarios, each filmed twice from three static cameras.
We use the Verified release with the official video-to-video protocol, where a method sees the first 3\,s of take 1 and the scenario description and predicts the next 5\,s.
We use one camera per scenario, the one with the largest motion area.
Besides the official Physics-IQ score, we report the average displacement error (ADE) of SAM~3 object tracks against the real future.

\subsection{State Accuracy and Visual Similarity}
\label{app:scene_metrics}
Visual similarity compares renders of the scene from its recovered camera path with the input frames, using Dual VQA (DV) and Latent Similarity (LS) from BVB~\cite{tang2026bvbbenchmarkingagenticvideo}.
State accuracy needs no image.
It compares the scene with the annotations behind VSI-Bench for object counts, object sizes, distances, and room area.

\subsection{Baselines and Implementation Details}
\label{app:impl}
The reasoning agent is Gemini 3.8 Flash~\cite{gemini38flash}, GPT-5.6 Luna~\cite{gpt56luna}, or Qwen3.8-27B~\cite{qwen27b}, the last one served locally with vLLM.
Write tasks use Gemini 3.8 Flash at medium reasoning effort.
A scene does not depend on the question, so it is built once and shared by all questions and reasoners.
The Video rows of~\cref{tab:vsi_full} give the same model the video and the question directly, and the Code (BVB) rows use scenes from the agent of BVB~\cite{tang2026bvbbenchmarkingagenticvideo}, read by our reasoning agent with the same tools, prompt, and budget.
The NVS baselines~\cite{yu2025trajectorycrafter,bai2025recammaster,chen2026reconstruct} get the same source video and target camera.
Perception and H3 run on RTX PRO 6000 GPUs.
Codification takes a median of 24 minutes for a room and 8 to 19 minutes for a write input.
With Gemini 3.8 Flash, a read question costs \$0.02 to \$0.07, and the 476 write tasks together cost \$97.

\section{Additional Results on State Reading}
\label{app:read}

\subsection{Full Results on VSI-Bench-Debiased}
\label{app:read_full}
\cref{tab:vsi_full} lists all eight question types.
Besides the three reasoners of the main paper, it includes Qwen3.6-27B, on which we also run SpatialClaw~\cite{cho2026spatialclaw}, and BVB scenes built with GPT-6 Astra.

\noindent\textbf{Where code input wins and loses.}
The gains sit in the two types that read metric 3D structure.
With our scenes, relative direction improves for every reasoner by 14.8 to 32.8 points, and absolute distance by 6.9 to 30.3.
Relative direction also improves with every BVB scene, so this gain comes from placing objects and observers in one 3D frame and not from a particular codifier.
Appearance order drops for every reasoner, by 4.1 to 14.5 points, and so does room size, by 3.0 to 17.4.
Route planning depends on the reasoner.
It rises by 20.2 and 11.1 points for Gemini and GPT and falls by 16.7 and 13.2 for the two Qwen models, which suggests that chaining several viewpoint queries needs a strong reasoner.

\noindent\textbf{Our scenes and BVB scenes.}
With the same codifier model (GPT-6.1 Terra), our scenes are on par with BVB's on average for three reasoners (within 0.2 points) and 1.7 points higher for Gemini.
The average hides a shift between question types.
For the three reasoners of the main paper, our scenes are better on object size (by 5.1 to 7.9 points) and absolute distance (8.4 to 9.6), which read metric geometry, and worse on object count (4.0 to 5.0).

\noindent\textbf{Comparison with a video agent.}
Unlike \model{}, which codifies a video once and then queries it with tools, SpatialClaw treats every question independently.
It calls spatial tools on the video inside a Python session and starts over for the next question.
On Qwen3.6-27B it has the best average (57.9, against 55.4 for video and 54.1 for our scenes) and is ahead of our scenes on five of eight types, most clearly route planning and object size.
Our scenes lead on relative direction by 26.0 points (75.7 vs.\ 49.7), where SpatialClaw falls below its own video baseline, and on object count by 7.1.
This accuracy comes at a price, and it stops at understanding.
The detailed cost comparison in \cref{tab:cost} reports 13.5 calls and 660.8K input tokens per question for SpatialClaw, versus 5.4 calls and 45.9K tokens for \model{}, or 2.5 times the calls and 14 times the input tokens.
A question costs \$0.180 instead of \$0.033, with accuracies of 66.4 and 65.3 on the cost-comparison questions. Including our \$1.52 build cost, the projected cumulative API cost becomes lower from the 11th query to the same scene (\cref{fig:cost-reuse}).
SpatialClaw also leaves no scene behind once a question is answered.
There is nothing to edit or render, so it cannot be applied to the write tasks (gray rows in \cref{tab:t5-bins,tab:t6-horizon}), whereas the scene that answers our questions is the same one we edit for NVS, GCP, and FP.

\input{tables/read_appendix}
\section{Additional Results on State Writing}
\label{app:write_results}

\subsection{Full Novel-View Synthesis Results and Visualization}
\label{app:nvs}

\cref{tab:t5-bins} adds SSIM, LPIPS, recall, and persistence to the main results.
\model{} is best on all 18 columns.
The appearance gap is moderate, with 1.2 to 2.0\,dB in PSNR and 0.07 to 0.09 in LPIPS over the best baseline.
The state gap is much larger.
Our mIoU is 0.22 to 0.25 and our recall 0.58 to 0.59, while no diffusion baseline exceeds 0.02 in mIoU or 0.09 in recall, and at 97$^\circ$ and 128$^\circ$ their mIoU is at most 0.004.
Our numbers are also nearly flat across the three angles (PSNR of 12.67 to 12.98), since a camera write leaves the scene untouched and a larger angle is not a harder edit.
Against the ceiling in the gray row, our recall reaches about 80\% and our mIoU about 40\%, which suggests that most objects are present in the new view and that the remaining error lies in their shape and pose.
PSNR stays low for every method, ours included, since our renders reproduce the layout but not the materials and lighting of the target.
SpatialClaw produces answers and no scene, so it has no output to evaluate here.

\begin{table*}
\centering
\small
\setlength{\tabcolsep}{2.5pt}
\renewcommand{\arraystretch}{1.15}

\caption{\textbf{Novel-View Synthesis results across view-change angles on Kubric-4D.}
We divide tasks into low-, medium-, and high-angle view changes
between the source and target cameras.
We extract object masks with SAM~3 and use Hungarian matching
to establish correspondences for mask-IoU evaluation.
Rec.\ and Pers.\ denote recall and persistence.
The gray row reports SAM~3 results on ground-truth videos.
Best and second-best results are \textbf{bold} and
\underline{underlined}.}
\label{tab:t5-bins}

\resizebox{\textwidth}{!}{%
\begin{tabular}{@{}l *{3}{rrrrrr}@{}}
\toprule
& \multicolumn{6}{c}{\textbf{Low} ($42^\circ$)}
& \multicolumn{6}{c}{\textbf{Medium} ($97^\circ$)}
& \multicolumn{6}{c}{\textbf{High} ($128^\circ$)} \\
\cmidrule(lr){2-7}
\cmidrule(lr){8-13}
\cmidrule(lr){14-19}

& \multicolumn{3}{c}{Appearance}
& \multicolumn{3}{c}{State}
& \multicolumn{3}{c}{Appearance}
& \multicolumn{3}{c}{State}
& \multicolumn{3}{c}{Appearance}
& \multicolumn{3}{c}{State} \\
\cmidrule(lr){2-4}\cmidrule(lr){5-7}
\cmidrule(lr){8-10}\cmidrule(lr){11-13}
\cmidrule(lr){14-16}\cmidrule(lr){17-19}

Method
& PSNR$\uparrow$ & SSIM$\uparrow$ & LPIPS$\downarrow$
& mIoU$\uparrow$ & Rec.$\uparrow$ & Pers.$\uparrow$
& PSNR$\uparrow$ & SSIM$\uparrow$ & LPIPS$\downarrow$
& mIoU$\uparrow$ & Rec.$\uparrow$ & Pers.$\uparrow$
& PSNR$\uparrow$ & SSIM$\uparrow$ & LPIPS$\downarrow$
& mIoU$\uparrow$ & Rec.$\uparrow$ & Pers.$\uparrow$ \\
\midrule

\textcolor{gray}{\textit{Ground Truth}}
& \textcolor{gray}{--} & \textcolor{gray}{--} & \textcolor{gray}{--}
& \textcolor{gray}{0.604} & \textcolor{gray}{0.731} & \textcolor{gray}{0.710}
& \textcolor{gray}{--} & \textcolor{gray}{--} & \textcolor{gray}{--}
& \textcolor{gray}{0.619} & \textcolor{gray}{0.751} & \textcolor{gray}{0.729}
& \textcolor{gray}{--} & \textcolor{gray}{--} & \textcolor{gray}{--}
& \textcolor{gray}{0.618} & \textcolor{gray}{0.749} & \textcolor{gray}{0.726} \\
\midrule

\rowcolor{gray!15}
SpatialClaw~\cite{cho2026spatialclaw} & \multicolumn{18}{c}{Not Applicable} \\ \midrule

\multicolumn{19}{c}{%
    \cellcolor{ourblue!10}\textit{Diffusion-based Methods}} \\
    
TrajectoryCrafter~\cite{yu2025trajectorycrafter}
& 10.88 & 0.310 & 0.681
& \underline{0.020} & 0.059 & 0.049
& 10.45 & 0.302 & 0.682
& \underline{0.004} & \underline{0.015} & \underline{0.014}
& 10.19 & 0.305 & 0.714
& \underline{0.004} & \underline{0.017} & \underline{0.015} \\

ReCamMaster~\cite{bai2025recammaster}
& \underline{11.68} & 0.397 & 0.683
& 0.016 & \underline{0.088} & \underline{0.062}
& 10.47 & 0.388 & 0.729
& 0.003 & \underline{0.015} & 0.011
& 9.34 & 0.372 & 0.789
& 0.001 & 0.005 & 0.003 \\

CogNVS~\cite{chen2026reconstruct}
& 11.49 & \underline{0.441} & \underline{0.644}
& 0.008 & 0.026 & 0.020
& \underline{11.04} & \underline{0.444} & \underline{0.648}
& 0.001 & 0.007 & 0.005
& \underline{10.69} & \underline{0.450} & \underline{0.663}
& 0.001 & 0.003 & 0.002 \\
\midrule

\model{} (Blender)
& \textbf{12.87} & \textbf{0.490} & \textbf{0.573}
& \textbf{0.254} & \textbf{0.594} & \textbf{0.519}
& \textbf{12.98} & \textbf{0.491} & \textbf{0.570}
& \textbf{0.230} & \textbf{0.580} & \textbf{0.510}
& \textbf{12.67} & \textbf{0.488} & \textbf{0.570}
& \textbf{0.224} & \textbf{0.583} & \textbf{0.500} \\
\bottomrule
\end{tabular}%
}
\end{table*}

\begin{figure}
    \centering
    \includegraphics[width=\linewidth]{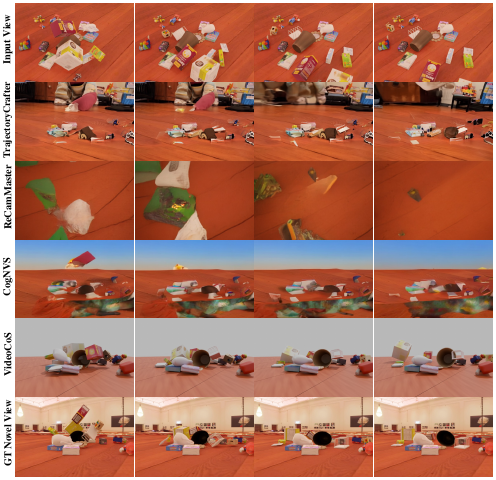}
    \caption{More qualitative results on novel-view synthesis.}
    \label{fig:supp-t5-scn02912-v0-v9}
\end{figure}

\begin{figure}
    \centering
    \includegraphics[width=\linewidth]{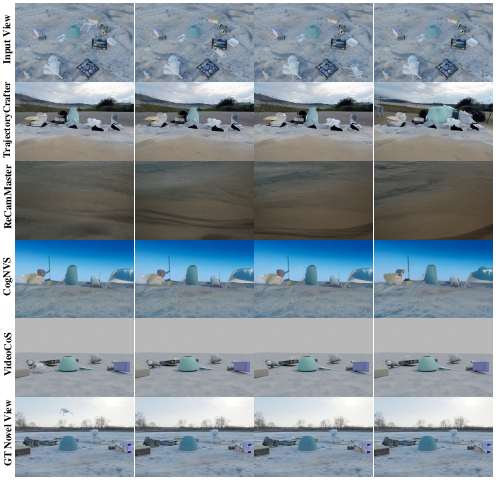}
    \caption{More qualitative results on novel-view synthesis.}
    \label{fig:supp-t5-scn02917-v0-v10}
\end{figure}

\begin{figure}
    \centering
    \includegraphics[width=\linewidth]{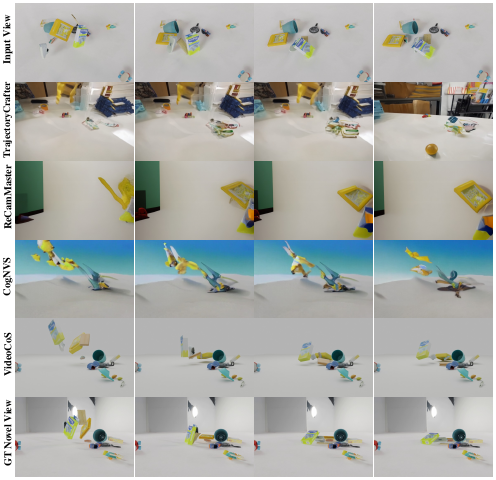}
    \caption{More qualitative results on novel-view synthesis.}
    \label{fig:supp-t5-scn02953-v0-v7}
\end{figure}

\begin{figure}
    \centering
    \includegraphics[width=\linewidth]{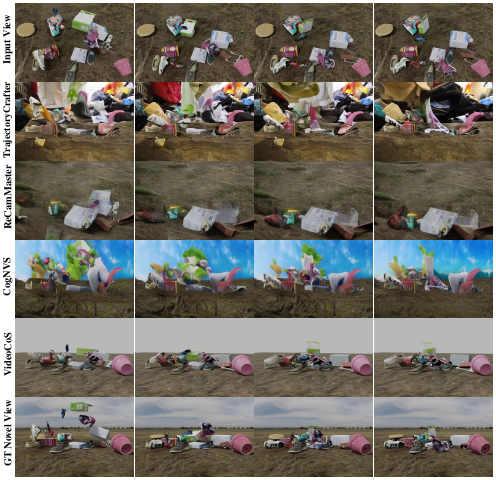}
    \caption{More qualitative results on novel-view synthesis.}
    \label{fig:supp-t5-scn02999-v0-v4}
\end{figure}

\begin{table*}[t]
\centering
\caption{\textbf{Goal-Conditioned Planning results by task horizon.}
Following Ego2Act, we split tasks by the median time humans take to complete
them: \textbf{Short} tasks fit within the 15\,s video budget (26 cases),
while \textbf{Long} tasks take longer (84 cases). \textbf{Total} includes all
110 cases.
The judge checks both goal completion and physical plausibility.
\textbf{Task} ($T$) covers starting the subgoal on the correct object (T1),
carrying out the main action (T2), and reaching the required end state (T3).
\textbf{Physical} ($P$) covers consistent object identities, counts, and
appearance (P1), a visible cause of motion (P2), realistic contact, support,
material behavior, and hand motion (P3), and a stable result afterwards (P4).
Each gate is scored independently as the percentage of yes judgments.
$T$ and $P$ average their respective gates; Final averages the per-video
score $\sqrt{T P}$.
Ours renders object motion without a hand, leading to low P2 scores.
Green/red mark gains/declines of the hybrid over MiniMax-H3 alone.}
\label{tab:t6-horizon}

\scriptsize
\setlength{\tabcolsep}{2pt}
\renewcommand{\arraystretch}{1.15}
\resizebox{\textwidth}{!}{%
\begin{tabular}{@{}l
    *{10}{c}@{\hspace{8pt}}
    *{10}{c}@{\hspace{8pt}}
    *{3}{c}@{}}
\toprule
\multirow{3}{*}{\textbf{Method}}
& \multicolumn{10}{c}{\textbf{Short} ($\leq 15$\,s)}
& \multicolumn{10}{c}{\textbf{Long} ($>15$\,s)}
& \multicolumn{3}{c}{\textbf{Total}} \\
\cmidrule(lr){2-11}\cmidrule(lr){12-21}\cmidrule(lr){22-24}

& \multicolumn{4}{c}{\textbf{Task}}
& \multicolumn{5}{c}{\textbf{Physical}}
& \multirow{2}{*}{\textbf{Final}}
& \multicolumn{4}{c}{\textbf{Task}}
& \multicolumn{5}{c}{\textbf{Physical}}
& \multirow{2}{*}{\textbf{Final}}
& \multirow{2}{*}{$\mathbf{T}$}
& \multirow{2}{*}{$\mathbf{P}$}
& \multirow{2}{*}{\textbf{Final}} \\
\cmidrule(lr){2-5}\cmidrule(lr){6-10}
\cmidrule(lr){12-15}\cmidrule(lr){16-20}

& \textbf{T1} & \textbf{T2} & \textbf{T3} & $\mathbf{T}$
& \textbf{P1} & \textbf{P2} & \textbf{P3} & \textbf{P4} & $\mathbf{P}$ &
& \textbf{T1} & \textbf{T2} & \textbf{T3} & $\mathbf{T}$
& \textbf{P1} & \textbf{P2} & \textbf{P3} & \textbf{P4} & $\mathbf{P}$ & & & & \\
\midrule

Seedance-2.0
& \textbf{92.7} & \textbf{89.9} & \textbf{80.8} & \textbf{87.8}
& \textbf{90.7} & \underline{95.7} & \textbf{88.3} & \underline{94.0}
& \textbf{92.2} & \textbf{89.6}
& \textbf{77.8} & \textbf{74.5} & \textbf{69.7} & \textbf{74.0}
& \textbf{85.9} & 91.6 & \textbf{84.1} & \underline{89.8}
& \textbf{87.8} & \textbf{79.0}
& \textbf{77.3} & \textbf{88.9} & \textbf{81.5} \\

Kling-v3-Pro
& 69.9 & 66.0 & 51.2 & 62.4
& 80.8 & \textbf{96.2} & \underline{82.7} & \textbf{94.9}
& \underline{88.6} & 71.5
& 60.5 & 60.9 & 51.5 & 57.6
& 84.6 & \textbf{93.4} & \underline{83.2} & \underline{89.8}
& \underline{87.7} & 68.8
& 58.8 & \underline{87.9} & 69.5 \\

Grok-1.5
& 73.3 & 74.6 & 65.1 & 71.0
& 69.3 & 90.5 & 78.7 & 86.3
& 81.2 & 72.8
& 63.2 & 64.9 & 55.0 & 61.1
& \underline{85.2} & 92.0 & 82.8 & \textbf{90.1}
& 87.5 & \underline{71.1}
& 63.4 & 86.0 & \underline{71.5} \\

Wan-2.7
& 72.7 & 76.3 & 67.9 & 72.3
& 80.6 & 92.9 & 80.6 & 85.8
& 85.0 & \underline{76.6}
& 64.2 & 61.5 & \underline{59.0} & 61.6
& 79.4 & \underline{92.3} & 78.9 & 82.4
& 83.3 & 68.9
& 64.1 & 83.7 & 70.6 \\

Cosmos-3
& 24.6 & 20.4 & 13.7 & 19.6
& 46.9 & 70.4 & 32.7 & 58.0
& 52.0 & 22.3
& 17.7 & 11.7 & 9.8 & 13.1
& 50.6 & 89.1 & 44.8 & 67.8
& 63.1 & 18.3
& 14.6 & 60.2 & 19.2 \\ 

MiniMax-H3
& 63.8 & 62.8 & 63.7 & 63.4
& 66.3 & 89.2 & 73.6 & 89.2
& 79.6 & 70.2
& 54.2 & 52.4 & 47.8 & 51.5
& 77.6 & 84.9 & 69.5 & 84.5
& 79.1 & 60.4
& 54.3 & 79.2 & 62.7 \\
\rowcolor{gray!15}
SpatialClaw & \multicolumn{23}{c}{Not Applicable} \\  \midrule

\multicolumn{24}{l}{\textit{\model{} plan in executable state, rendered by}} \\
\quad Blender
& \underline{82.5} & \underline{87.9} & 73.7 & \underline{81.3}
& \underline{89.4} & 0.0 & 38.1 & 86.3
& 53.5 & 62.0
& \underline{68.0} & \underline{67.1} & 57.5 & \underline{64.2}
& 76.3 & 3.2 & 38.8 & 77.3
& 48.9 & 49.6
& \underline{68.2} & 50.0 & 52.5 \\
\quad Blender + H3
& \gain{76.6} & \gain{79.3} & \gain{\underline{79.0}} & \gain{78.3}
& \gain{73.3} & \loss{72.0} & \loss{53.1} & \loss{71.5}
& \loss{67.5} & \loss{67.7}
& \gain{55.8} & \gain{62.0} & \gain{58.6} & \gain{58.8}
& \loss{71.0} & \loss{73.4} & \loss{58.7} & \loss{77.0}
& \loss{70.0} & \loss{57.9}
& \gain{63.5} & \loss{69.4} & \loss{60.1} \\

\bottomrule
\end{tabular}%
}
\end{table*}

\subsection{Full Goal-Conditioned Planning Results and Visualization}
\label{app:gcp}
\cref{tab:t6-horizon} breaks the two scores into their gates.
On the task gates, the Blender plan ranks second to Seedance-2.0 in T1 and T2 on both short and long tasks, third in short-task T3, and fourth in long-task T3.
The low physical score comes from two gates.
P2 asks for a visible cause of motion and is 0.0 and 3.2, since objects move without a hand, and P3, which covers contact and hand motion, is 38.1 and 38.8.
The other two gates are in the range of the video generators on short tasks (89.4 for P1 and 86.3 for P4).
Re-rendering with full H3 improves P2 and P3, raising P2 to 72.0 and 73.4 and P3 to 53.1 and 58.7 on short and long tasks, respectively.
It also loses part of the plan, as T falls from 81.3 to 78.3 on short tasks and from 64.2 to 58.8 on long ones.
The hybrid still completes more tasks than full H3 alone (T of 63.5 vs.\ 54.3 in total), but its physical and final scores remain lower (69.4 vs.\ 79.2 and 60.1 vs.\ 62.7).
Every method drops from short to long tasks, ours by 17.1 points in T and Seedance-2.0 by 13.8.
As in NVS, SpatialClaw is not applicable.

\begin{figure}
    \centering
    \includegraphics[width=\linewidth]{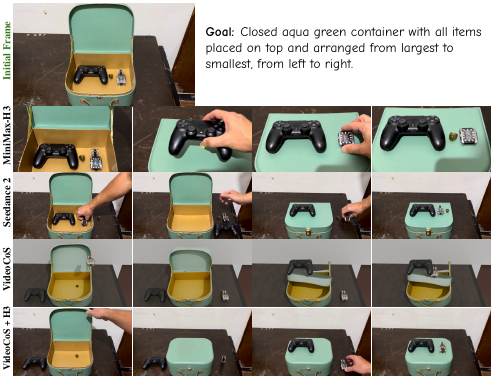}
    \caption{More qualitative results on goal-conditioned planning.}
    \label{fig:supp-t6-controller-watch-ring}
\end{figure}

\begin{figure}
    \centering
    \includegraphics[width=\linewidth]{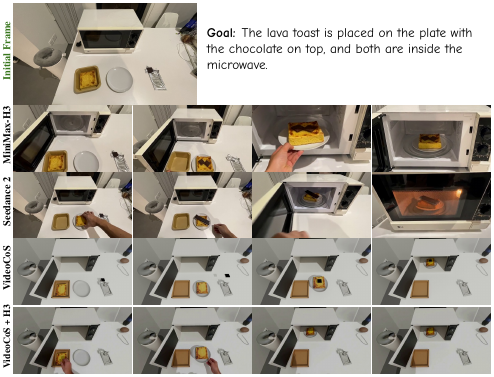}
    \caption{More qualitative results on goal-conditioned planning.}
    \label{fig:supp-t6-lava-toast-microwave}
\end{figure}

\begin{figure}
    \centering
    \includegraphics[width=\linewidth]{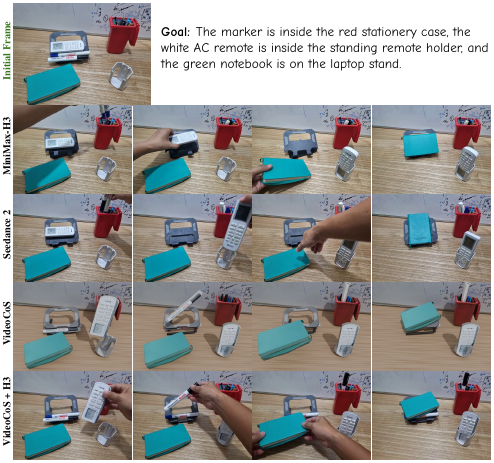}
    \caption{More qualitative results on goal-conditioned planning.}
    \label{fig:supp-t6-remote-marker-stand}
\end{figure}

\begin{figure}
    \centering
    \includegraphics[width=\linewidth]{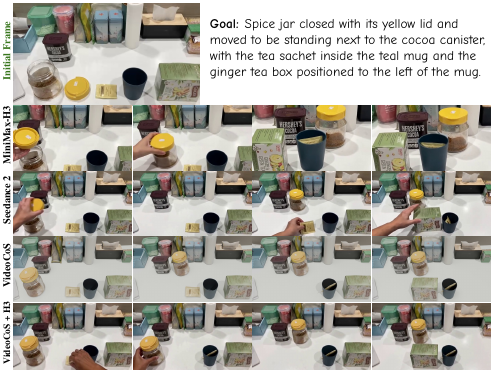}
    \caption{More qualitative results on goal-conditioned planning.}
    \label{fig:supp-t6-pantry-jar-sachet}
\end{figure}

\subsection{Future Prediction Visualization}
\label{app:fp}
The examples in this section are rigid scenarios (a ball on a ramp, a ball rolling on glass, spaced dominoes, and a ball rolling behind a box), the regime where the simulated plan follows the real future most closely (rigid ADE of 11.1 in \cref{tab:t7-main}).

\begin{figure}
    \centering
    \includegraphics[width=\linewidth]{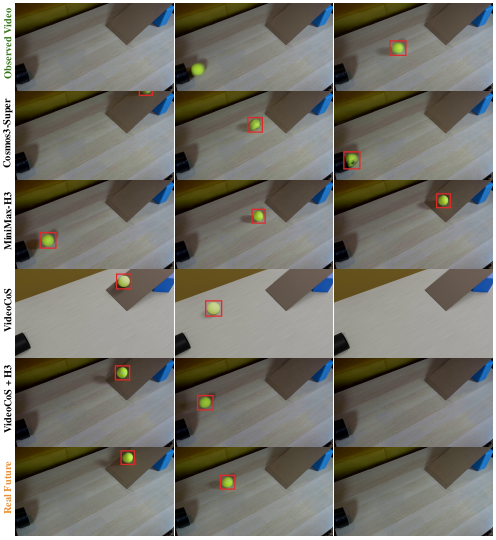}
    \caption{More qualitative results on future prediction.}
    \label{fig:supp-t7-0019-left-ball-ramp}
\end{figure}

\begin{figure}
    \centering
    \includegraphics[width=\linewidth]{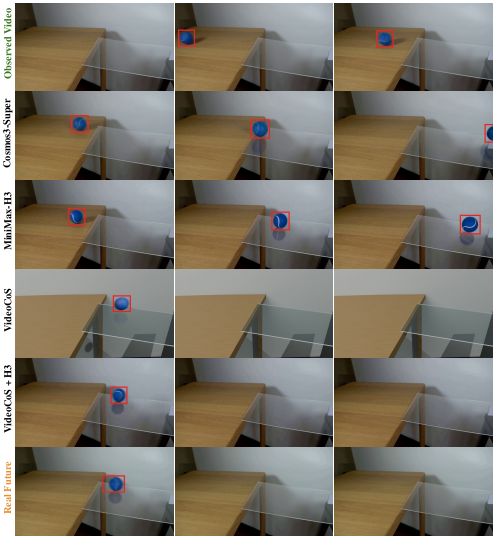}
    \caption{More qualitative results on future prediction.}
    \label{fig:supp-t7-0027-right-ball-rolls-on-glass}
\end{figure}

\begin{figure}
    \centering
    \includegraphics[width=\linewidth]{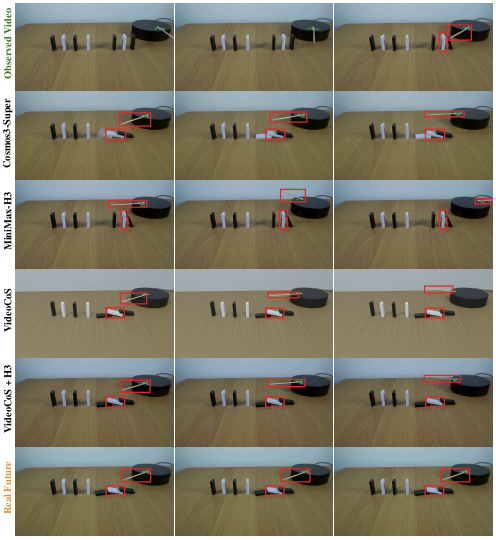}
    \caption{More qualitative results on future prediction.}
    \label{fig:supp-t7-0050-center-dominos-with-space}
\end{figure}

\begin{figure}
    \centering
    \includegraphics[width=\linewidth]{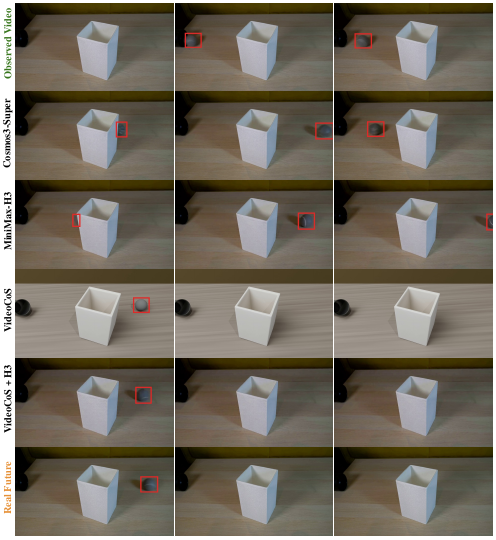}
    \caption{More qualitative results on future prediction.}
    \label{fig:supp-t7-0146-center-roll-behind-box}
\end{figure}

\section{More Analysis}

\noindent\textbf{What evidence should the codification agent get?}
We fix the codification agent (Codex CLI with GPT-6.1 Terra) and vary the perception evidence it receives (\cref{tab:codify_evidence}).
\emph{Frames and cameras} gives the input frames and cameras and no other evidence.
\emph{Fused objects} gives one 3D box per object (the ``3D boxes, identities'' row of \cref{tab:supp_evidence}), obtained by lifting SAM~3 tracks with DA3 depth and merging them across frames.
\emph{Unfused layers} gives the same perception before fusion (per-frame depth, cameras, and SAM~3 masks with track ids and names) and lets the agent decide which tracks form an object.
The setting we use for write tasks adds two aids to the unfused layers.
A \emph{contact} query casts camera rays through the bottom edge of an object's mask and intersects them with the surface the object rests on, which locates where the object stands without using depth.
\emph{Mask overlays} draw selected tracks on the frames, so that the agent can check what each track covers.

Fused objects are quick to copy but carry perception errors into the scene.
DA3 depth covers only the surfaces that face the camera, so the boxes are truncated and overlap, which gives 46 interpenetrations spread over seven of the ten samples.
Unfused layers remove most of them (5), but the object center error is the highest (0.44\,m), consistent with lifted mask points lying on the visible surface of an object and not at its center.
The contact query fixes the position of each object on its supporting surface independently of depth.
The center error drops to 0.29\,m, the 3D IoU matches that of the fused objects (0.33 vs.\ 0.32), and one interpenetration remains, for two to three more minutes of codification per sample.
This is a pilot with one run per setting.
A later run of the final setting on the same four Kubric-4D scenes gives 0.35\,m and 0.28 IoU, so we do not interpret geometry differences below about 0.06\,m.

\begin{table}
\centering
\footnotesize
\setlength{\tabcolsep}{4pt}
\renewcommand{\arraystretch}{1.05}
\caption{\textbf{Evidence given to the codification agent.} Pilot on ten write-task samples. Err.\ is the object center error and IoU the 3D IoU, both on four Kubric-4D scenes. Interp.\ is the total number of interpenetrations, and Min.\ is the median codification time per sample. The last row is the setting we use for write tasks.}
\label{tab:codify_evidence}
\begin{tabular}{@{}l cccc@{}}
\toprule
\textbf{Evidence} & Err.\,(m)$\downarrow$ & IoU$\uparrow$ & Interp.$\downarrow$ & Min. \\
\midrule
Frames and cameras & 0.36 & 0.23 & 10 & 4.2 \\
Fused objects & 0.40 & 0.32 & 46 & 4.2 \\
Unfused layers & 0.44 & 0.23 & 5 & 4.8 \\
\rowcolor{oursbg}\quad + contact, mask overlays & \textbf{0.29} & \textbf{0.33} & \textbf{1} & 7.0 \\
\bottomrule
\end{tabular}
\end{table}

\subsection{Quality of Codified Scenes}
\label{app:scene_quality}

\begin{table}
\centering
\small
\setlength{\tabcolsep}{2.5pt}
\caption{\textbf{Quality and accuracy of the codified scenes for the 274 VSI-Bench-Debiased rooms.}
QA accuracy uses Gemini 3.8 Flash. 
See \cref{app:scene_metrics} for metric details.}
\resizebox{\columnwidth}{!}{%
\begin{tabular}{@{}l ccc ccccc c@{}}
\toprule
\multirow{2}{*}{\textbf{Method}}
& \multicolumn{3}{c}{\textbf{Visual Similarity}}
& \multicolumn{5}{c}{\textbf{State Accuracy}}
& \multirow{2}{*}{\textbf{QA Acc.}} \\
\cmidrule(lr){2-4}\cmidrule(lr){5-9}
& \textbf{DV} & \textbf{LS} & \textbf{Overall}
& \textbf{Count} & \textbf{Size} & \textbf{Dist.}
& \textbf{Area} & \textbf{Avg.} & \\
\midrule
BVB (Terra)
& 44.1 & 88.1 & 64.2
& \textbf{66.2} & 44.5 & 42.7 & 53.0 & 52.2
& 63.8 \\
BVB (Astra)
& 56.6 & 88.5 & 71.7
& 62.7 & 40.8 & 39.5 & \textbf{62.2} & 52.1
& 62.6 \\ \midrule
\rowcolor{oursblue} Ours (Terra)
& \textbf{60.3} & \textbf{89.0} & \textbf{73.9}
& 64.9 & \textbf{53.6} & \textbf{49.0} & 61.6 & \textbf{57.9}
& \textbf{65.5} \\
\bottomrule
\end{tabular}%
}
\label{tab:vsi-state}
\end{table}

As listed in~\cref{tab:vsi-state}, our scenes are more accurate in state than BVB's (57.9 vs.\ 52.2) while being only slightly closer in appearance (73.9 vs.\ 71.7).
The full read results show the same pattern at the level of answers.
Our scenes lead where the answer is a metric value, on object size and absolute distance, and trail on object count.

\begin{figure}
    \centering
    \includegraphics[width=\linewidth]{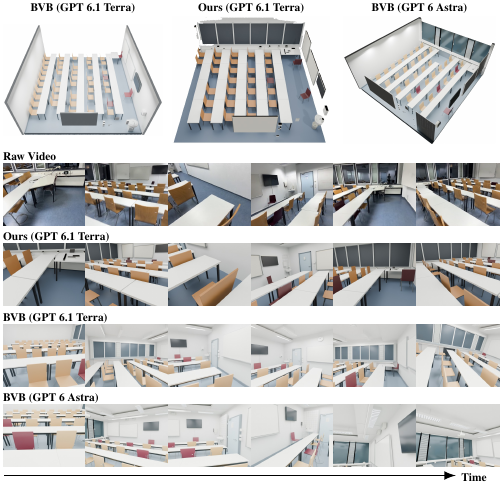}
    \caption{Top: Scene reconstructions from different codification
methods/agents. Bottom: Input video frames and corresponding
rendered frames. \model{} more accurately reproduces camera
poses, visual details, and object state.}
    \label{fig:supp-21d970d8de}
\end{figure}

\begin{figure}
    \centering
    \includegraphics[width=\linewidth]{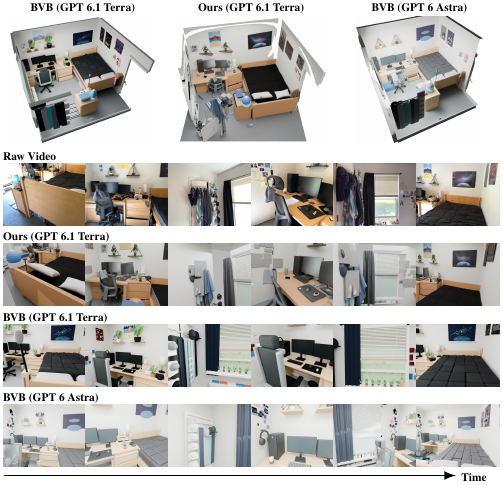}
    \caption{Top: Scene reconstructions from different codification
methods/agents. Bottom: Input video frames and corresponding
rendered frames. \model{} more accurately reproduces camera
poses, visual details, and object state.}
    \label{fig:supp-c49a8c6cff}
\end{figure}

\begin{figure}
    \centering
    \includegraphics[width=\linewidth]{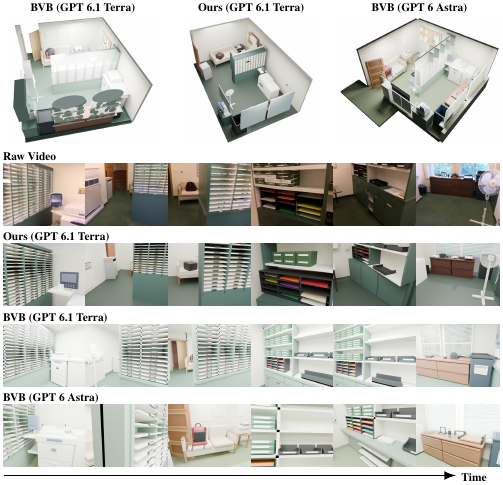}
    \caption{Top: Scene reconstructions from different codification
methods/agents. Bottom: Input video frames and corresponding
rendered frames. \model{} more accurately reproduces camera
poses, visual details, and object state.}
    \label{fig:supp-scene0593-00}
\end{figure}

\begin{figure}
    \centering
    \includegraphics[width=\linewidth]{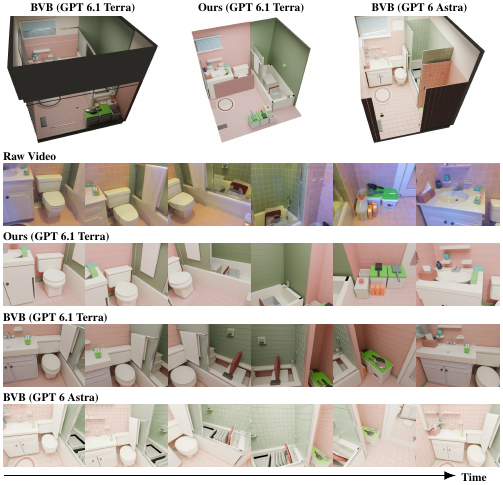}
    \caption{Top: Scene reconstructions from different codification
methods/agents. Bottom: Input video frames and corresponding
rendered frames. \model{} more accurately reproduces camera
poses, visual details, and object state.}
    \label{fig:supp-scene0693-00}
\end{figure}

\subsection{Cost Breakdown}
\label{app:cost}

\cref{tab:cost} reports the measured build and per-question costs underlying \cref{fig:cost-reuse}.
For $N$ queries to the same scene, the curves use
\begin{align*}
C_{\mathrm{ours}}(N) &= 1.52 + 0.033N,\\
C_{\mathrm{SpatialClaw}}(N) &= 0.180N,\\
C_{\mathrm{video}}(N) &= 0.011N.
\end{align*}
Costs are in US dollars and assume one codification per scene.
These are linear projections from measured average costs over three scenes and 90 questions.
The API-cost crossover with SpatialClaw is $N=1.52/(0.180-0.033)\approx10.34$, so \model{} becomes cheaper from the 11th query.
At 18 and 48 queries, its projected costs are 65.2\% and 35.9\% of SpatialClaw's, respectively.

\begin{table}[t]
\centering
\footnotesize
\setlength{\tabcolsep}{3pt}
\caption{\textbf{Build and per-question cost}, averaged over three VSI-Bench-Debiased scenes (90 questions) with Gemini 3.8 Flash answering. API charges and GPU usage are reported separately.}
\renewcommand{\arraystretch}{1.05}
\resizebox{\linewidth}{!}{%
\begin{tabular}{@{}l cc cccc@{}}
\toprule
& \multicolumn{2}{c}{\textbf{Build (per scene)}} & \multicolumn{4}{c}{\textbf{Answer (per question)}} \\
\cmidrule(lr){2-3}\cmidrule(l){4-7}
\textbf{Method} & \textbf{API (\$)} & \textbf{GPU (h)} & \textbf{Calls} & \textbf{In / Out tok. (K)} & \textbf{API (\$)} & \textbf{GPU (min)} \\
\midrule
Gemini 3.8 Flash (video) & -- & -- & 1 & 8.0 / 1.3 & 0.011 & 0 \\
\quad + SpatialClaw~\cite{cho2026spatialclaw} & -- & -- & 13.5 & 660.8 / 14.5 & 0.180 & $\le$1.2 \\
\rowcolor{oursblue}\quad + \model{} & 1.52 & 2.17 & 5.4 & 45.9 / 2.5 & 0.033 & 0 \\
\bottomrule
\end{tabular}}

\label{tab:cost}
\end{table}

The API curves exclude GPU compute charges.
Our scene construction uses 2.17 GPU hours per scene, while SpatialClaw uses up to 1.2 GPU minutes per question in this comparison.
Lower API cost therefore does not establish lower GPU usage or lower end-to-end cost.
The 65.3 versus 66.4 accuracy comparison refers to the cost-comparison questions; the 65.5 accuracy in \cref{fig:analysis} refers to the full reading evaluation.
These costs quantify repeated read queries, not state edits or video generation.

\subsection{Limitations and Future Work}
\label{app:future}

\noindent\textbf{Limitations.}
A codified scene only contains what a rigid-body Blender program can express.
It has objects with joints, a camera, and lights, but no model of hands, fluids, deformable materials, or fire, which is why \model{} trails video generators on physical and visual realism (Sec.~4.3).
Codification takes a frontier coding agent, up to 30 minutes of agent time, and about two GPU hours of perception for a room video ($\leq$10 minutes).
Finally, our evaluation covers indoor rooms, synthetic rigid scenes, table-top manipulation, and short physics clips.
Outdoor scenes, moving people, and much longer videos are untested, and codification and the write tasks each rely on a single model.

\noindent\textbf{Future work.}
The most direct extension is a hybrid state, in which the executable scene keeps geometry and rigid motion while learned components supply what it cannot simulate, such as hands, non-rigid dynamics, and photorealistic appearance.
Our MiniMax-H3 variant is a first step, but it still trades control for realism.
A second direction is to let the agent decide when to trust the state.
Attaching a confidence to each object and checking doubtful values against the video would target the cases where code input loses to video input, such as counts and appearance order.
A third is cheaper codification, for example by distilling the build--check--fix loop into a smaller model, or by updating a scene incrementally as a long video comes in instead of rebuilding it.

%% file: tables/read_appendix.tex
\begin{table*}[t]
\centering
\footnotesize
\setlength{\tabcolsep}{3pt}
\renewcommand{\arraystretch}{1.05}
\caption{
\textbf{State querying on VSI-Bench-Debiased.}
Code representations are constructed using BVB or our codification harness.
Astra and Terra denote GPT-6 Astra and GPT-6.1 Terra, respectively.
Green and red indicate improvements and declines relative to the
corresponding video baseline.
}
\label{tab:vsi_full}

\begin{tabular}{@{}lll *{9}{c}@{}}
\toprule
& & &
\textbf{Entity}
& \multicolumn{4}{c}{\textbf{Geometry}}
& \multicolumn{2}{c}{\textbf{Viewpoint}}
& \textbf{History} & \\
\cmidrule(lr){4-4}
\cmidrule(lr){5-8}
\cmidrule(lr){9-10}
\cmidrule(lr){11-11}

\textbf{Reasoner}
& \textbf{State Representation}
& \textbf{Codifier}
& Count & Size & Room & Abs.\ Dist.
& Rel.\ Dist. & Rel.\ Dir. & Route & Order
& \textbf{Avg.} \\
\midrule

Qwen3.6-27B & Video & --
& 46.2 & 58.9 & \textbf{58.9} & 38.1
& 60.2 & 53.8 & 50.0 & \textbf{77.0} & 55.4 \\

\quad + SpatialClaw~\cite{cho2026spatialclaw} & Video & --
& \gain{51.2} & \gain{\textbf{64.5}} & \loss{51.8}
& \gain{\textbf{53.8}} & \loss{56.8} & \loss{49.7}
& \gain{\textbf{60.7}} & \loss{75.0} & \gain{\textbf{57.9}} \\

\quad + \model{} & Code (BVB) & GPT-6 Astra
& \gain{55.2} & \loss{50.9} & \loss{54.9}
& \gain{45.9} & \gain{\textbf{65.2}} & \gain{72.3}
& \loss{43.0} & \loss{61.9} & \gain{56.2} \\

\quad + \model{} & Code (BVB) & GPT-6.1 Terra
& \gain{53.7} & \loss{58.2} & \loss{41.7}
& \gain{52.3} & \gain{60.4} & \gain{72.4}
& \loss{34.3} & \loss{57.9} & \loss{53.9} \\

\rowcolor{oursblue}
\quad + \model{} & Code (Ours) & GPT-6.1 Terra
& \gain{\textbf{58.3}} & \loss{52.2} & \loss{41.5}
& \gain{45.0} & \gain{60.6} & \gain{\textbf{75.7}}
& \loss{36.8} & \loss{62.5} & \loss{54.1} \\
\midrule

Qwen3.8-27B & Video & --
& 51.7 & \textbf{62.4} & 52.2 & 40.4
& \textbf{69.3} & 57.5 & \textbf{60.5}
& \textbf{74.5} & \textbf{58.6} \\

\quad + \model{}& Code (BVB) & GPT-6 Astra
& \gain{57.2} & \loss{51.9} & \gain{\textbf{54.6}}
& \gain{46.4} & \loss{66.1} & \gain{70.5}
& \loss{50.8} & \loss{61.9} & \loss{57.4} \\

\quad + \model{}& Code (BVB) & GPT-6.1 Terra
& \gain{\textbf{58.1}} & \loss{54.8} & \loss{44.7}
& \gain{48.0} & \loss{60.9} & \gain{75.5}
& \loss{52.6} & \loss{64.4} & \loss{57.4} \\

\rowcolor{oursblue}
\quad + \model{}& Code (Ours) & GPT-6.1 Terra
& \gain{54.1} & \loss{61.5} & \loss{43.7}
& \gain{\textbf{56.4}} & \loss{62.6} & \gain{\textbf{76.5}}
& \loss{43.8} & \loss{61.9} & \loss{57.6} \\
\midrule

GPT-5.6 Luna & Video & --
& 49.4 & 61.8 & 54.0 & 25.9
& 59.4 & 40.0 & 41.2 & \textbf{68.6} & 50.0 \\

\quad + \model{} & Code (BVB) & GPT-6 Astra
& \gain{53.2} & \loss{53.6} & \gain{\textbf{58.1}}
& \gain{45.7} & \gain{\textbf{67.1}} & \gain{69.1}
& \gain{59.6} & \loss{61.3} & \gain{58.5} \\

\quad + \model{} & Code (BVB) & GPT-6.1 Terra
& \gain{\textbf{59.0}} & \loss{56.8} & \loss{51.3}
& \gain{47.4} & \gain{60.0} & \gain{\textbf{72.9}}
& \gain{\textbf{64.9}} & \loss{63.8} & \gain{59.5} \\

\rowcolor{oursblue}
\quad + \model{} & Code (Ours) & GPT-6.1 Terra
& \gain{54.0} & \gain{\textbf{61.9}} & \loss{50.8}
& \gain{\textbf{56.2}} & \gain{63.6} & \gain{72.8}
& \gain{52.3} & \loss{64.5} & \gain{\textbf{59.6}} \\
\midrule

Gemini 3.8 Flash & Video & --
& \textbf{59.6} & \textbf{67.4} & 57.2 & 42.9
& \textbf{73.5} & 66.8 & 59.6 & \textbf{77.7} & 63.1 \\

\quad + \model{} & Code (BVB) & GPT-6 Astra
& \loss{55.8} & \loss{53.6} & \gain{\textbf{59.7}}
& \gain{47.3} & \loss{71.9} & \gain{71.4}
& \gain{77.2} & \loss{64.2} & \loss{62.6} \\

\quad + \model{} & Code (BVB) & GPT-6.1 Terra
& \loss{58.5} & \loss{56.5} & \loss{52.9}
& \gain{48.7} & \loss{65.8} & \gain{77.9}
& \gain{\textbf{80.7}} & \loss{69.2} & \gain{63.8} \\

\rowcolor{oursblue}
\quad + \model{} & Code (Ours) & GPT-6.1 Terra
& \loss{53.7} & \loss{64.4} & \loss{54.2}
& \gain{\textbf{58.3}} & \loss{67.4} & \gain{\textbf{81.6}}
& \gain{79.8} & \loss{64.5} & \gain{\textbf{65.5}} \\
\bottomrule
\end{tabular}
\end{table*}